\documentclass{article}

    \usepackage[final]{neurips_2022}

\usepackage{graphicx}
\usepackage[utf8]{inputenc} 
\usepackage[T1]{fontenc}    
\usepackage{hyperref}       
\usepackage{url}            
\usepackage{booktabs}       
\usepackage{amsfonts}       
\usepackage{nicefrac}       
\usepackage{microtype}      
\usepackage{xcolor}         
\usepackage{algorithm}  
\usepackage{algorithmic} 
\usepackage{tikz}
\usepackage{comment}
\usepackage{amsmath,amssymb} 
\usepackage{color}
\usepackage{subfigure}
\usepackage{natbib}
\setcitestyle{numbers,square}
\title{Stimulative Training of Residual Networks: A Social Psychology Perspective of Loafing}

\author{Peng Ye\textsuperscript{1}$^\dagger$~,~Shengji Tang\textsuperscript{1}$^\dagger$~,~Baopu Li\textsuperscript{2}~,~Tao Chen\textsuperscript{1}\thanks{Corresponding Author (eetchen@fudan.edu.cn).~~~$^\dagger$Equal Contribution.}~,~Wanli Ouyang\textsuperscript{3}\\
\textsuperscript{1}School of Information Science and Technology, Fudan University, \textsuperscript{2}Oracle Health and AI, USA, \\
\textsuperscript{3}The University of Sydney, SenseTime Computer Vision Group, Australia, and Shanghai AI Lab\\
{\tt\small yepeng20@fudan.edu.cn}}

\begin{document}
\bibliographystyle{unsrt} 

\maketitle

\begin{abstract}
Residual networks have shown great success and become indispensable in today’s deep models. In this work, we aim to re-investigate the training process of residual networks from a novel social psychology perspective of loafing, and further propose a new training strategy to strengthen the performance of residual networks.
As residual networks can be viewed as ensembles of relatively shallow networks (i.e., \textit{unraveled view}) in prior works, we also start from such view and consider that the final performance of a residual network is co-determined by a group of sub-networks. 
Inspired by the social loafing problem of social psychology, we find that residual networks invariably suffer from similar problem, where sub-networks in a residual network are prone to exert less effort when working as part of the group compared to working alone. We define this previously overlooked problem as \textit{network loafing}. As social loafing will ultimately cause the low individual productivity and the reduced overall performance, network loafing will also hinder the performance of a given residual network and its sub-networks. Referring to the solutions of social psychology, we propose \textit{stimulative training}, 
which randomly samples a residual sub-network and calculates the KL-divergence loss between the sampled sub-network and the given residual network, to act as extra supervision for sub-networks and make the overall goal consistent. Comprehensive empirical results and theoretical analyses verify that stimulative training can well handle the loafing problem, and improve the performance of a residual network by improving the performance of its sub-networks. The code is available at \href{https://github.com/Sunshine-Ye/NIPS22-ST}{https://github.com/Sunshine-Ye/NIPS22-ST}. 
\end{abstract}

\section{Introduction}

Since ResNet~\cite{he2016deep} wins the first place at the ILSVRC-2015 competition, simple-but-effective residual connections are applied in various deep networks, such as CNN, MLP, and transformer. To explore the secrets behind the success of residual networks, numerous studies have been proposed. He et. al~\cite{he2016deep} exploit the residual structure to avoid the performance degradation of deep networks. Further, He et. al~\cite{he2016identity} consider that such a structure can transfer any low level features to high level layers in forward propagation and directly transmit the gradients from deep to shallow layers in backward propagation. Balduzzi et. al~\cite{balduzzi2017shattered} find that residual networks can alleviate the shattered gradients problem that gradients resemble white noise. In addition, Veit et. al~\cite{veit2016residual} experimentally verify that residual networks can be seen as a collection of numerous networks of different lengths, namely \textit{unraveled view}. Following this view, Sun et. al~\cite{sun2022low} further attribute the success of residual networks to shallow sub-networks, which may correspond to the low-degree term when regarding the neural network as a polynomial function. Since the unraveled view is supported both experimentally and theoretically, we further investigate some interesting mechanism behind residual networks based on this inspiring work. Specifically, we treat a residual network as the ensemble of relatively shallow sub-networks and consider its final performance to be co-determined by a group of sub-networks.
\begin{figure}
\vspace{-6mm}
\centering
\includegraphics[height=5.5cm]{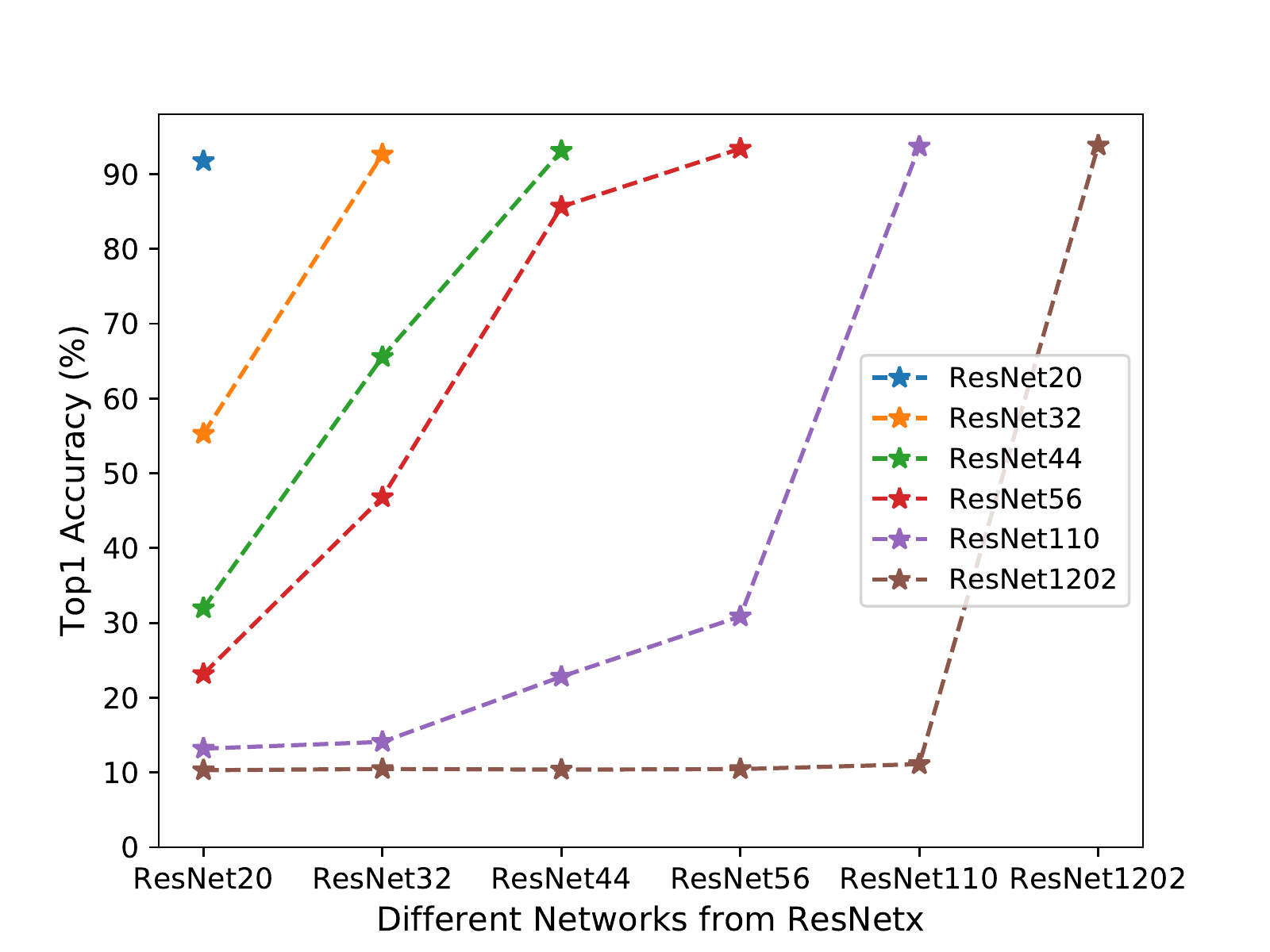}
\vspace{-3mm}
\caption{Different residual networks invariably suffer from the problem of network loafing, and deeper residual network tends to have more serious loafing problem. All these networks are trained on CIFAR10 dataset. The horizontal axis means the sampled different sub-networks from ResNetx.} 
\label{fig:acc_resnetx}
\vspace{-5mm}
\end{figure}
\par In social psychology, working in a group is always a tricky thing. Compared with performing tasks alone, group members tend to make less efforts when working as part of a group, which is defined as the social loafing problem~\cite{ingham1974ringelmann,petty1977effects}. Moreover, social psychology researches find that increasing the group size may aggravate social loafing for the decrease of individual visibility~\cite{williams1991social,george1995asymmetrical}. Inspired by these, we find that the ensemble-like networks formed by residual connections also have a similar problem and behavior. As shown in Fig. 1, we can see that, different residual networks invariably suffer from the loafing problem, that is, the sub-networks working in a given residual network are prone to exert degraded performances than these sub-networks working individually. For example, a sub-network ResNet32 in the ResNet56 only has a 46.80\% Top1 accuracy, much lower than ResNet32 trained individually with accuracy of 92.63\%. Moreover, the loafing problem of deeper residual networks is inclined to be more severe than that of shallower ones, that is, the same sub-network in deeper residual networks constantly presents inferior performance than that in shallower residual networks. For example, ResNet20 within ResNet32 has a 55.28\% Top1 accuracy, while ResNet20 within ResNet56 (deeper than ResNet 32) only has a 23.17\% Top1 accuracy.  Such problems have not been addressed in the literature so far as we know. Hereafter, we define this previously overlooked problem as \textit{network loafing}.\footnote{\textit{Network loafing} is just a loose analogy to describe a behavior in neural networks that has no strong connection with biology.} As social psychology researches show that social loafing will ultimately cause low productivity of each individual and the collective
~\cite{latane1979many,simms2014social}, we consider that network loafing may also hinder the performance of given residual network and all of its sub-networks.
\par In social psychology, there are two commonly used solutions for preventing social loafing within groups: 1) establishing individual accountability by increasing the \emph{individual supervision} and 2) making tasks cooperative by setting up the overall goal~\cite{williams1991social,george1995asymmetrical}. Inspired by this, we propose a novel training strategy for improving residual networks, namely \textit{stimulative training}. In details, for each mini-batch during stimulative training, besides the main loss of the given residual network in conventional training, we will randomly sample a residual sub-network (individual supervision) and calculate the KL-divergence loss between the sampled sub-network and the given residual network (consistent overall goal). 
This simple yet effective training strategy can relieve the loafing problem of residual networks, by strengthening the individual supervision of sub-networks and making the goals of residual sub-networks and the given residual network more consistent.
\par Comprehensive empirical analyses verify that stimulative training can solve the loafing problem of residual networks effectively and efficiently, thus improve both the performance of a given residual network and all of its residual sub-networks by a large margin. Furthermore, we theoretically show the connection of the proposed stimulative training strategy and the improved performance of a given residual network and all of its residual sub-networks. Besides, experiments on various benchmark datasets using various residual networks demonstrate the effectiveness of the proposed training strategy. The contributions of our work can be summarized as the following:
\begin{itemize}
\vspace{-7pt}
\item We understand residual networks from a social psychology perspective, and find that different residual networks invariably suffer from the problem of network loafing.
\vspace{-2pt}
\item We improve residual networks from a social psychology perspective, and propose a simple-but-effective stimulative training strategy to improve the performance of the given residual network and all of its sub-networks.
\vspace{-2pt}
\item Comprehensive empirical and theoretical analysis verify that stimulative training can well solve the loafing problem of residual networks.
\vspace{-7pt}
\end{itemize}


\section{Related Works}
\subsection{Unraveled View} \vspace{-1mm}
As one pioneer work to investigate residual networks,~\cite{veit2016residual} 
experimentally shows 
that residual networks can be seen as a collection of numerous networks of different length, namely unraveled view. Subsequently,~\cite{sun2022low} follows this view and further attributes the success of residual networks to shallow sub-networks first. Besides,~\cite{sun2022low} considers the neural network as a polynomial function and corresponds shallow sub-network to low-degree term to explain the working mechanism of residual networks. Similar to~\cite{veit2016residual} and~\cite{sun2022low}, this paper also investigates residual networks from the unraveled view. Differently, inspired by social psychology, we further reveal the loafing problem of residual networks under the unraveled view. Besides, we propose a novel stimulative training method to relieve this problem and further unleash the potential of residual networks.

\subsection{Knowledge Distillation}  \vspace{-1mm}
As a classical method, knowledge distillation~\cite{hinton2015distilling, qiu2022better} transfers the knowledge from a teacher network to a student network via approximating the logits~\cite{hinton2015distilling,kim2018paraphrasing,mirzadeh2020improved} or features~\cite{romero2014fitnets,huang2017like,heo2019knowledge, xu2018pad} output. To avoid the huge cost of training a high performance teacher, some works abandon the naive teacher-student framework, like mutual distillation~\cite{zhang2018deep} making group of students learn from each other online, and self distillation~\cite{zhang2019your} transferring knowledge from deep layers to shallow layers. Generally, all these distillation methods need to introduce additional networks or structures, and employ fixed teacher-student pairs. As a comparison, our method does not require any additional network or structure, and the student network is a randomly sampled sub-network of a network. Besides, our method is essentially designed to address the loafing problem of residual networks, which is different from knowledge distillation that aims to obtain a compact network with acceptable accuracy.
\subsection{One-shot NAS} \vspace{-1mm}
One-shot NAS is an important branch of neural architecture search (NAS)~\cite{ye2022b,chen2021bn,ye2022efficient,li2020improving,ye2022beta,guo2022neural}. Along this direction,~\cite{cai2019once} trains an once-for-all (OFA) network with progressive shrinking and knowledge distillation to support kinds of architectural settings. Following this work, BigNAS~\cite{yu2020bignas} introduces several technologies to train a high-quality single-stage model, whose child models can be directly deployed without extra retraining or post-processing steps. Both OFA and BigNAS aim at simultaneously training and searching various networks with different resolutions, depths, widths and operations. Differently, the proposed method aims at improving a given residual network, thus can be seamlessly applied to the searched model of NAS. As OFA and BigNAS are not designed to solve the loafing problem, their sampling space and supervision signal are also different with the proposed method. More importantly, the social-psychology-inspired problem of network loafing may explain why OFA and BigNAS work.
\subsection{Stochastic Depth} \vspace{-1mm}
As a regularization technique, Stochastic Depth~\cite{huang2016deep} randomly disables the convolution layers of residual blocks, to reduce training time and test error substantially. In Stochastic Depth, the reduction in test error is attributed to strengthening gradients of earlier layers and the implicit ensemble of numerous sub-networks of different depths. In fact, its improved performance can be also interpreted as relieving the network loafing problem defined in this work. The theoretical analysis in this work can be also applied to Stochastic Depth. Besides, for better solving the loafing problem, our method samples sub-networks with ordered depth, and uses an additional KL-divergence loss to provide a more achievable target and make the output of a given network and its sub-networks more consistent.

\section{Stimulative Training}
\subsection{Motivation} \vspace{-1mm}
Social loafing is a social psychology phenomenon that individuals lower their productivity when working in a group. Based on the novel perspective that a residual network behaves like an ensemble network~\cite{veit2016residual}, we find various residual networks invariably exhibit loafing-like behaviors as shown in Fig.~\ref{fig:acc_resnetx} and Fig.~\ref{fig:diff_training_strategy}, which we define as network loafing.
As social loafing is ``a kind of social disease" that will harm the individual and collective productivity~\cite{latane1979many}, we consider network loafing may also hinder the performance of a residual network and its sub-networks. To alleviate network loafing, it is intuitive to learn from social psychology. There are two common methods for solving social-loafing problem in sociology, namely, establishing individual accountability (i.e., increasing the individual supervision) and making tasks cooperative (i.e., setting up the overall goal)~\cite{williams1991social,george1995asymmetrical}. In order to increase individual supervision, we sample sub-networks in the whole network and provide extra supervision to train each sub-network sufficiently. For the overall goal, we adopt KL divergence loss to constrain the output of sub-networks not far from that of the whole network,
which aims at reducing the performance gap and driving the sampled sub-networks to develop cooperatively.
\begin{figure}[t]
\vspace{-6mm}
\centering
\includegraphics[height=6.4cm]{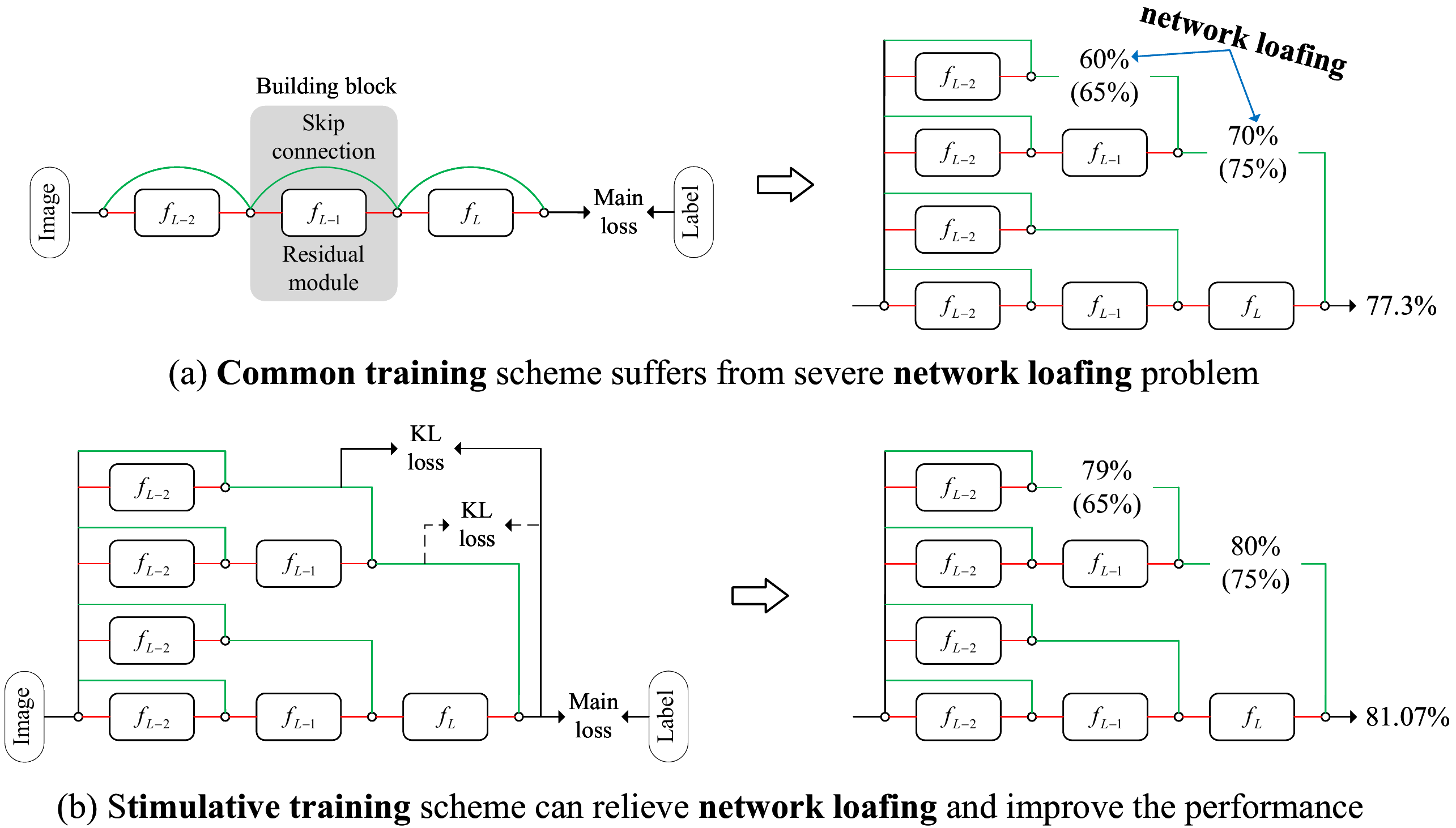}
\vspace{-3mm}
\caption{Illustration of common and stimulative training schemes. Stimulative training can relieve the network loafing problem, and improve the performance of a given residual network (from 77.3\% to 81.07\%) and all of its sub-networks (e.g., from 60\% to 79\%). 65\% and 75\% denote the individual performance of each sub-network.}
\label{fig:UT scheme}
\vspace{-4mm}
\end{figure}

\subsection{Training Algorithm} \vspace{-1mm}
In this subsection, we briefly illustrate the working scheme of the proposed stimulative training strategy, and show its difference from the common training method. As shown in Fig.~\ref{fig:UT scheme},
common training only focuses on optimization of main network, thus suffers from severe network loafing, that is, sub-networks lower their performance when working in an ensemble. For example, as shown in Fig.~\ref{fig:UT scheme}(a) sub-networks within the residual network only have an accuracy of 60\% and 70\%, much lower than their individual accuracy of 65\% and 75\%. As a comparison, stimulative training optimizes the main network and meanwhile uses it to provide extra supervision for a sampled sub-network at each training iteration, thus well handles the loafing problem. In the test procedure, our method can adopt the main network or any sub-network as the inference model, thus requiring the same or lower memory and inference cost compared with a given residual network. 
\par Formally, for a given residual network to be optimized, we define the main network as $D_m$ and the sub-network as $D_s$. All the sub-networks share weights with the main network, and make up the sampling space $\Theta = \left \{ D_s|D_s = \pi (D_m) \right \}$, where $\pi$ is sampling operator, usually random sampling.
In the training process, we randomly sample a sub-network at each iteration to ensure the whole sampling space can be fully explored. To make the training more efficient and effective, we define a new sampling space obeying ordered residual sampling rule to be discussed in Section~\ref{sec:ors}. Denoting $\theta_D{}_m$ and $\theta_D{}_s$ as weights of the main network and the sampled network respectively, $x$ as the mini-batch training sample and $y$ as its label, $\mathcal{Z}$ as the output of network, the total loss of stimulative training is computed as
\begin{align}
\label{fomulation:ut loss} \mathcal{L}_{mt} = \underbrace{ CE(\mathcal{Z}(\theta_{D_m},x ),y)}_{\rm{main\ supervision}} + \underbrace{\lambda KL(\mathcal{Z}(\theta_{D_m},x),\mathcal{Z}(\theta_{D_s} ,x))}_{\rm {extra\ common \ direction \ supervision}},D_s \in \Theta 
\end{align}
where the $CE$ and $KL$ are the standard cross entropy loss and KL divergence loss.  $\lambda$ is the balancing coefficient, and $CE$ loss denotes the main supervision for the main network.
$KL$ loss provides an extra supervision for the sampled sub-network, and guarantees all sub-networks to be optimized in the same direction with the main network, which can be considered as setting a common goal for the ``ensemble network”. 
We use the standard stochastic gradient descent for updating the model, denoted as
$
\theta_{D_m}^{t+1} = \theta_{D_m}^{t} - \eta \frac{\partial \mathcal L_{ut}}{\partial \theta_{D_m}} 
$
, where $\eta$ is the learning rate. Thanks to the weight-sharing between sub-networks and the main network, we only need to update $\theta_{D_m}$ once for each iteration to optimize both the $\theta_{D_m}$ and $\theta_{D_s}$. The pseudo code is shown in Appendix C.1.
\begin{figure}[t]
\centering
\includegraphics[height=6.8cm]{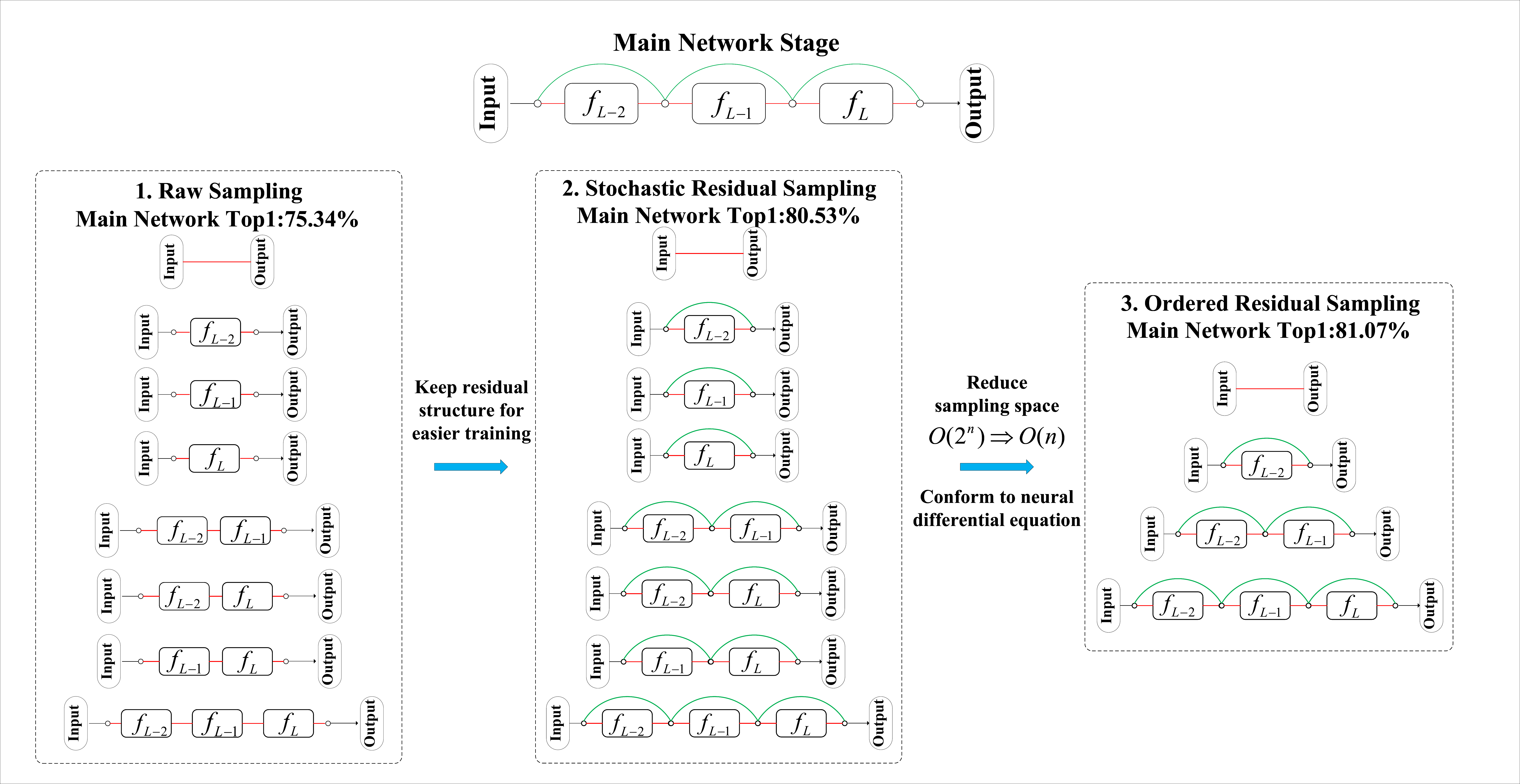}
\vspace{-3mm}
\caption{Illustration of ordered residual unraveling rule. Compared with two other sampling strategies, ordered residual sampling can facilitate training and reduce the sampling space to get better main network performance. }
\label{fig:sampling way}
\vspace{-4mm}
\end{figure}
\subsection{Ordered Residual Sampling} \vspace{-1mm} \label{sec:ors}
Network loafing is based on the novel view that residual networks can be seen as the ensemble of numerous different sub-networks\cite{veit2016residual}. It is intuitive to adopt raw unraveled view to design sampling space, namely raw sampling. However, there exist two problems in raw sampling. First, 
it introduces a mass of single branch sub-networks which are hard to optimize, as shown in the left sub-figure of Fig.~\ref{fig:sampling way}. 
Second, with the growth of network depth, the size of sampling space increases exponentially. For a residual stage of $n$ blocks, there are $2^n$ sub-networks, which constitutes a too large space to explore. Under limited computation resources, it is difficult to provide sufficient supervision to each sub-network, causing insufficient stimulative training.

Therefore, in order to facilitate stimulative training, we re-design two different sampling ways: stochastic residual sampling and ordered residual sampling. Fig.~\ref{fig:sampling way} shows a three-block residual network as a simple illustration for three sampling ways. For raw sampling, we obey the raw definition of~\cite{veit2016residual}, that means 
each sub-network is composed of stacked convolution layers. For stochastic residual sampling, we keep the basic structure of residual blocks and skip some blocks randomly. For ordered residual sampling, we also keep the residual structure but skip blocks orderly, (i.e., always skip the last several blocks). We implement these sampling methods for MobileNetV3~\cite{howard2019searching} on CIFAR-100. Experimental results show that ordered residual sampling performs better than other two methods (by 5.77\% and 0.58\% respectively). To explain the superiority of ordered residual sampling, we should pay attention to keeping residual structure and ordered sampling. For keeping residual structure, it can drive the network training process easier and improve the final performance~\cite{he2016deep}.
For ordered sampling, it can noticeably reduce the size of sampling space (from $O(2^n)$ to $O(n)$), making it possible to train each sub-network sufficiently. Besides, all deep networks with residual connections can be approximated by neural differential equations~\cite{chen2018neural}, which treat discrete network layers as continuous and regard the layer output as the result of adding an integral to the layer input. From this perspective, ordered sampling will maintain the continuity of residual networks. 
Because of these outstanding features, we will adopt ordered residual sampling in the following experiments.                      
\section{Empirical Analysis}
In this section, we conduct extensive experiments to analyze and compare the proposed stimulative training and the common training strategy in detail. We first illustrate the loafing problem of residual networks with different training strategies. We further demonstrate statistical characteristics of the performance of the trained residual network and all of its sub-networks. Then, we show the performance drop when destructing trained residual networks at test time, including deleting individual layers, deleting more layers and permuting layers. Besides, we also record the KL distance between a given residual network and all of its sub-networks in the whole training process. For the convenience of analysis, we always use a NAS-searched model MobileNetV3~\cite{howard2019searching} as the residual network and train it on CIFAR100, unless otherwise specified. The detailed experimental settings can be found in Appendix C.2.

\subsection{Experiment 1: Checking the loafing problem of residual networks with different training strategies} \vspace{-1mm}
\begin{figure}[t]
\vspace{-4mm}
  \centering
  \subfigure[6 sub-networks]{
    \label{fig:subfig:6} 
    \includegraphics[width=0.32\linewidth]{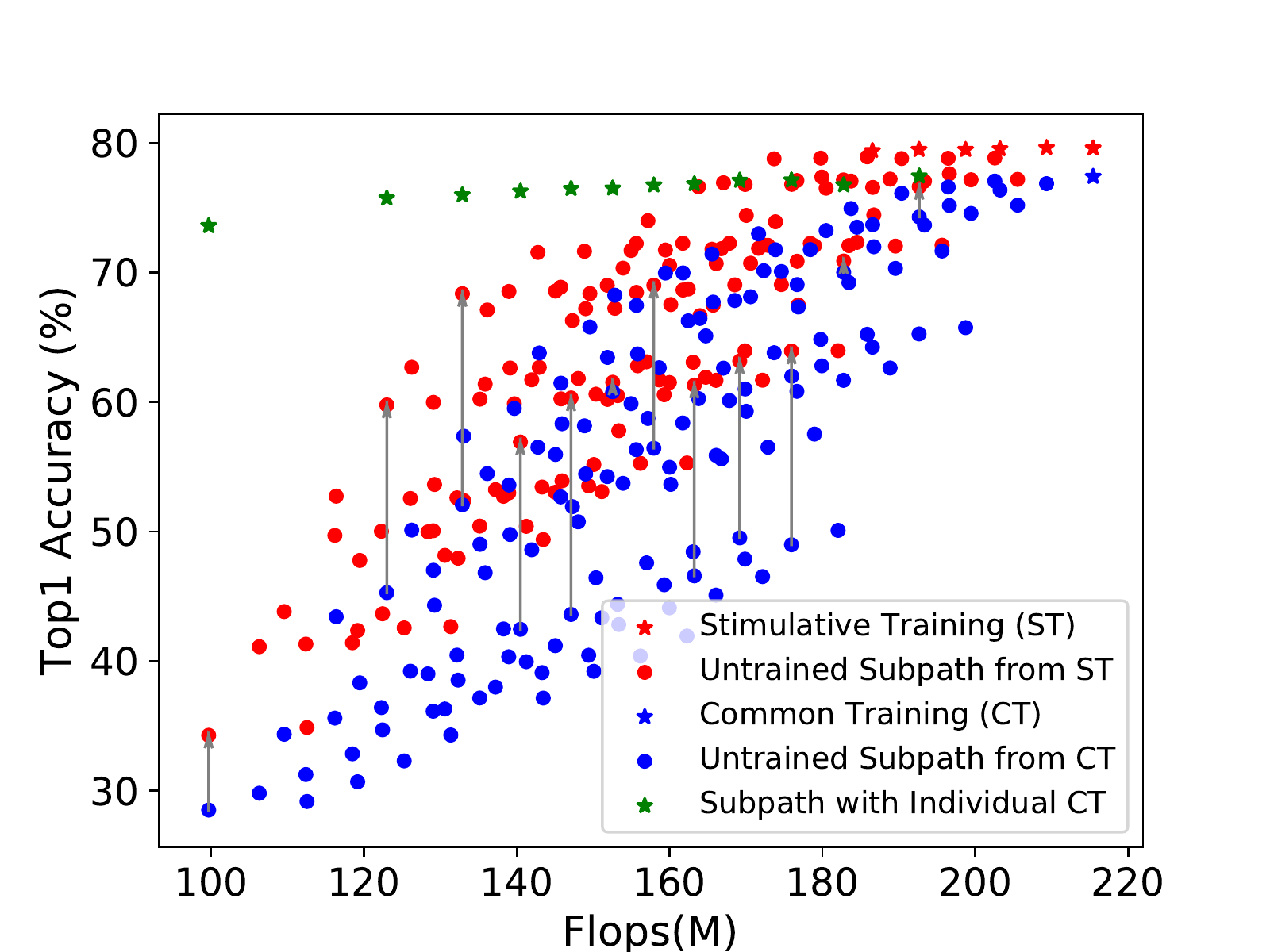}}
  \subfigure[48 sub-networks]{
    \label{fig:subfig:48} 
    \includegraphics[width=0.32\linewidth]{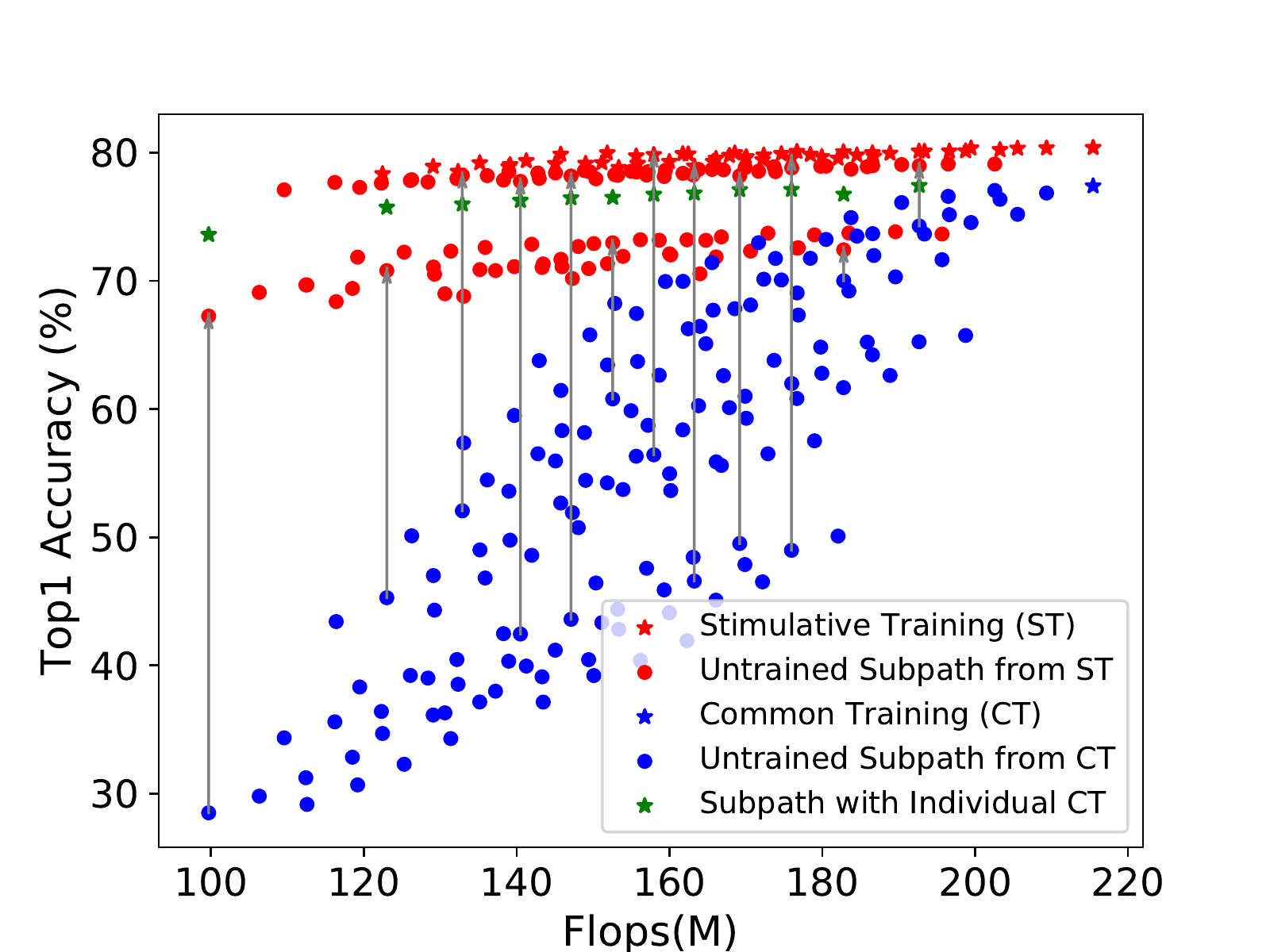}}
  \subfigure[144 sub-networks]{
    \label{fig:subfig:144} 
    \includegraphics[width=0.32\linewidth]{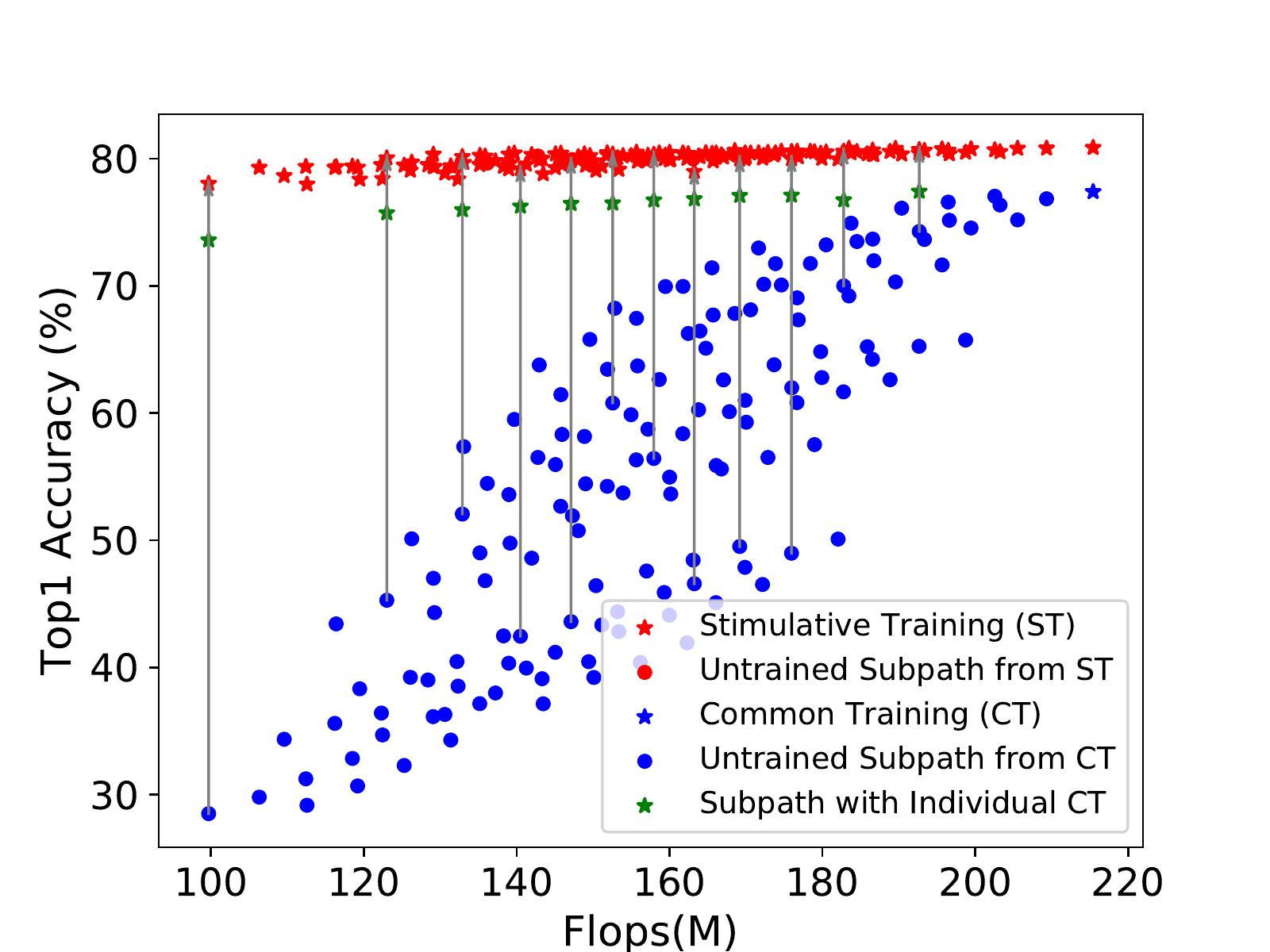}}
  \vspace{-3mm}
  \caption{Stimulative training using 6, 48, and 144 sub-networks. With more sub-networks involved, the network loafing problem could be relieved better, and the performance of the residual network and all of its sub-networks could be improved more significantly. We train MobileNetV3 with different schemes on CIFAR100 dataset.}
  \label{fig:diff_training_strategy} 
\vspace{-4mm}
\end{figure}

\begin{figure}[t]
  \centering
  \subfigure[Maximum Value]{
    \label{fig:subfig:max} 
    \includegraphics[width=0.32\linewidth]{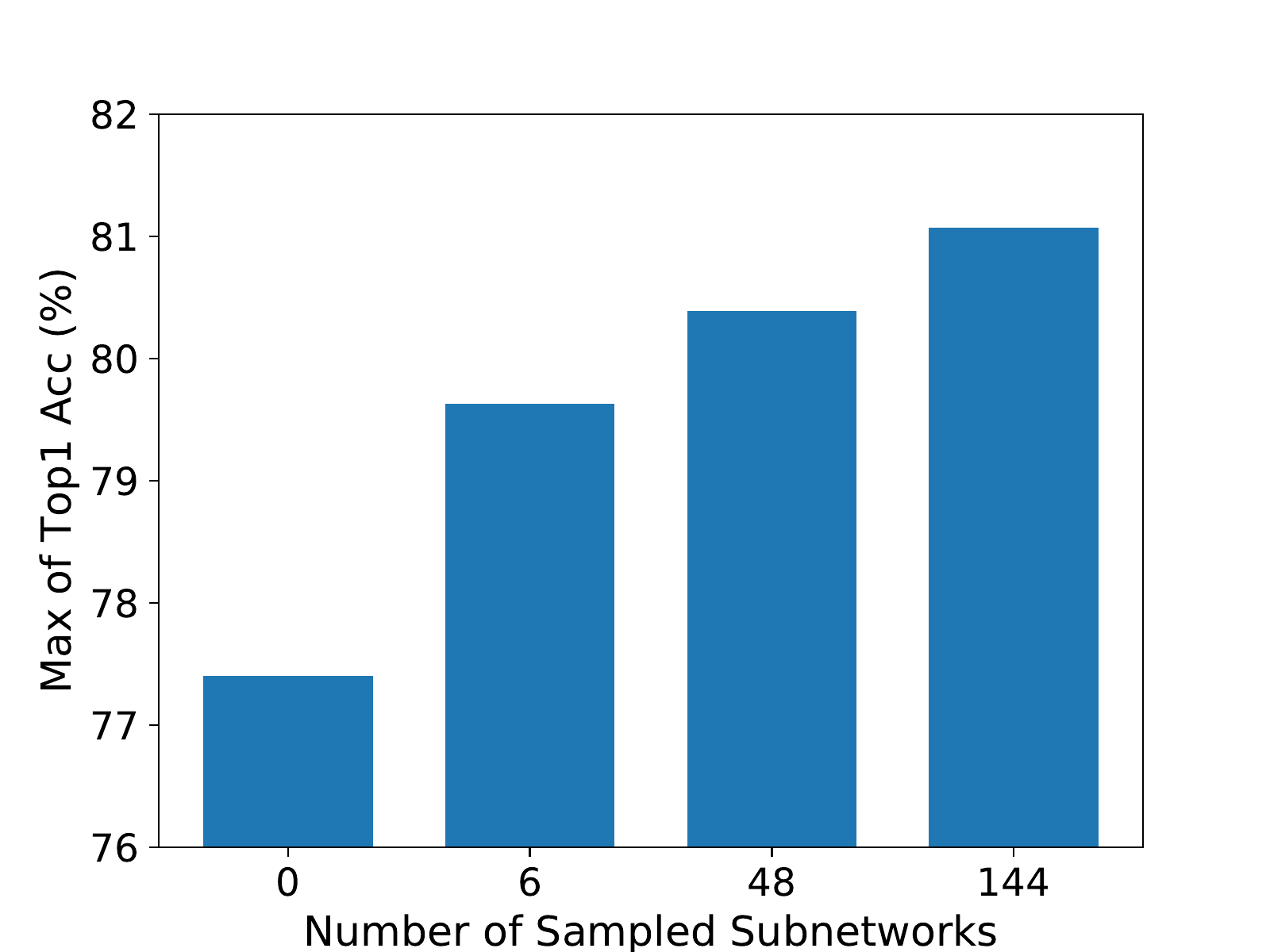}}
  \subfigure[Mean Value]{
    \label{fig:subfig:mean} 
    \includegraphics[width=0.32\linewidth]{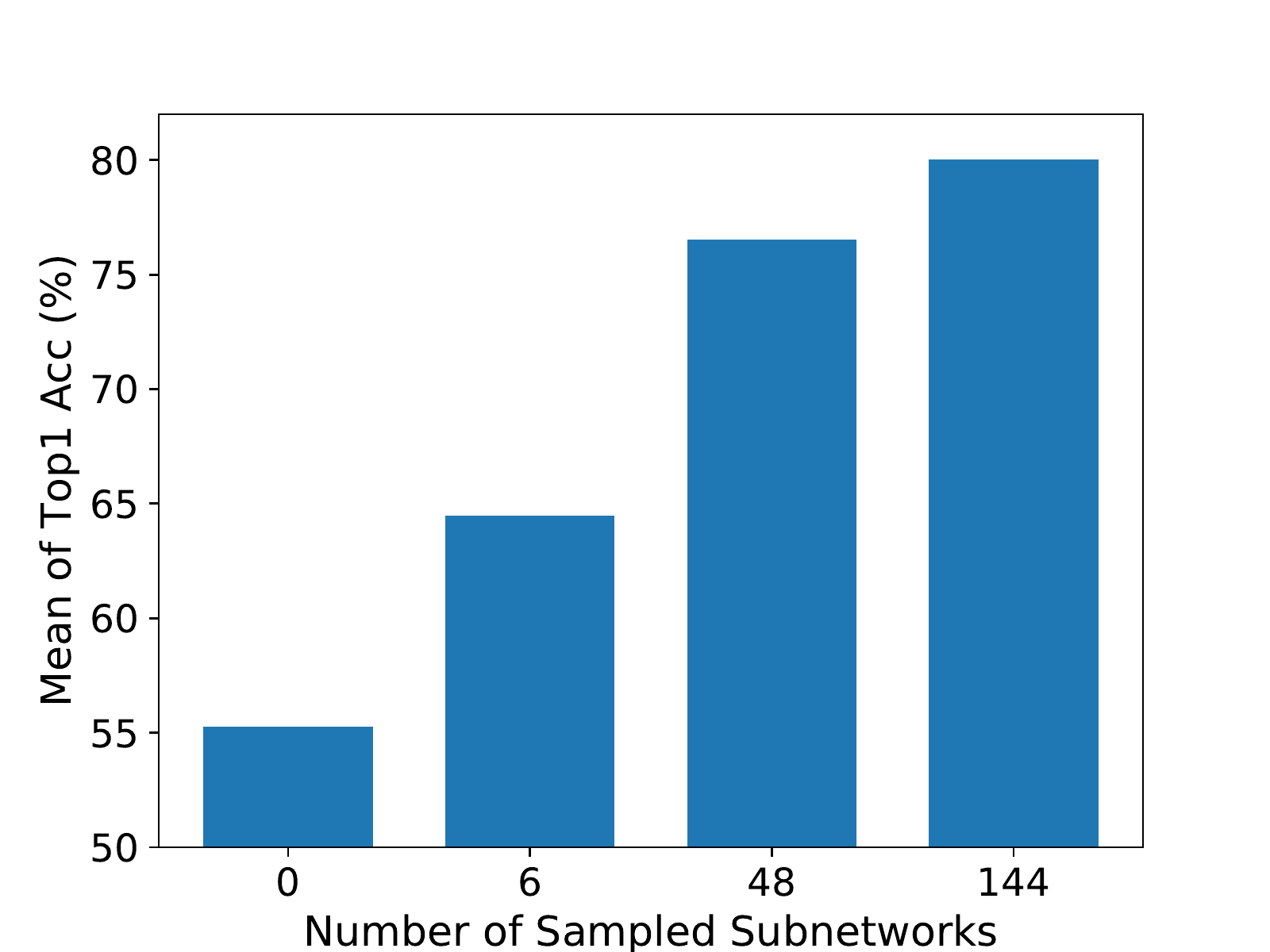}}
  \subfigure[Standard Deviation]{
    \label{fig:subfig:std} 
    \includegraphics[width=0.32\linewidth]{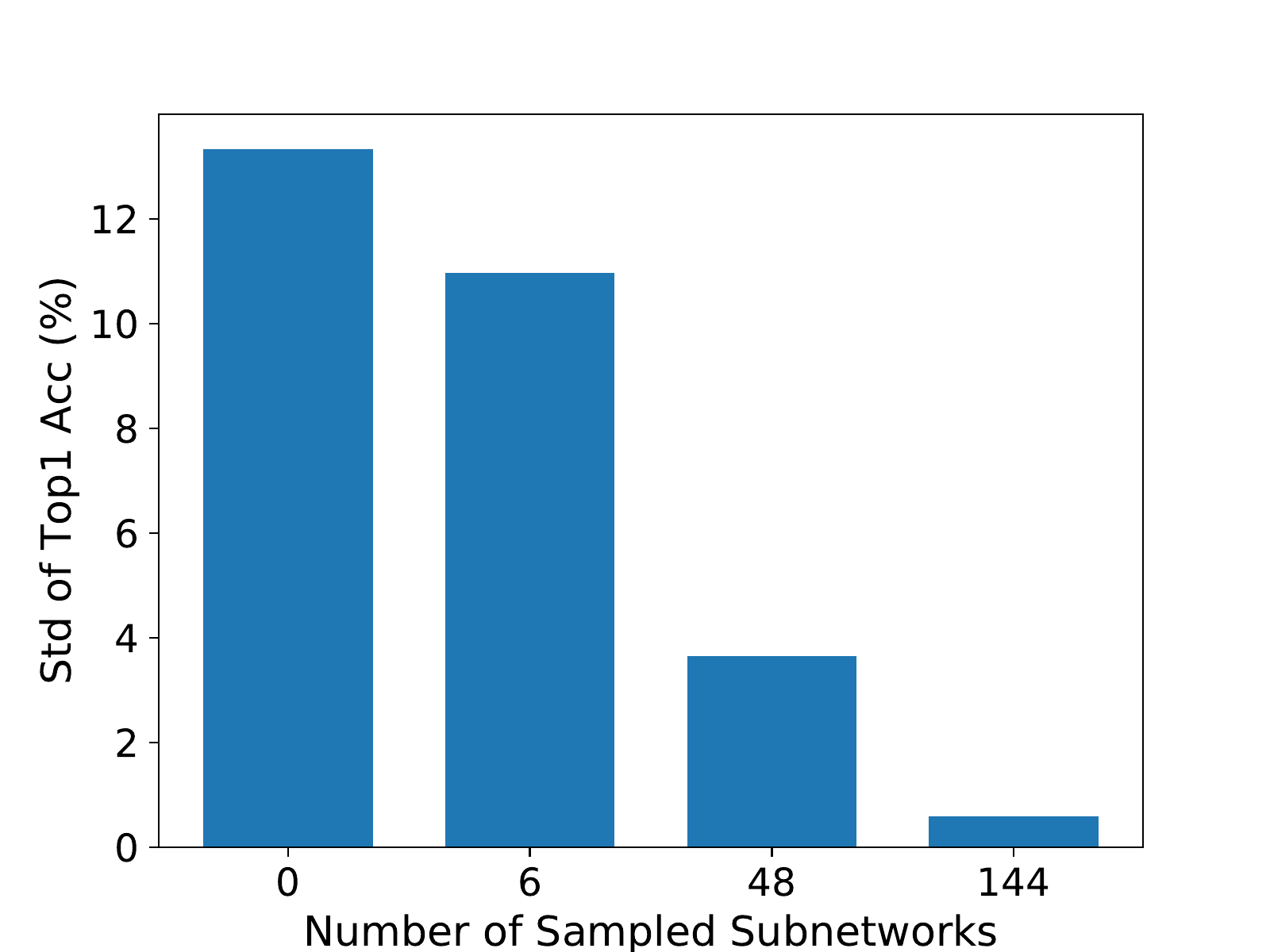}}
  \vspace{-3mm}
  \caption{(a) Maximum, (b) mean value and (c) standard deviation when training MobileNetV3 with different strategies on CIFAR100 dataset. Sampling 0 sub-networks denotes the common training strategy. Stimulative training with more sub-networks can provide higher maximum, mean and lower standard deviation.}
  \label{fig:statistical} 
\vspace{-4mm}
\end{figure}
First, we find that the NAS-searched residual network MobileNetV3 still suffers from the loafing problem. As shown in Fig.~\ref{fig:diff_training_strategy}, when applying the common training strategy to MobileNetV3, the sub-networks from MobileNetV3 (blue solid circle) have much lower accuracies than the same sub-networks individually trained (green five-pointed star), and the performance gap increases when the size of the sub-network decreases. Second, we show that the proposed stimulative training strategy can well handle the loafing problem. As shown in Fig.~\ref{fig:subfig:144}, when applying the stimulative training strategy to MobileNetV3, the sub-networks from MobileNetV3 (red five-pointed star) have 
even better performance than the same sub-networks individually trained (green five-pointed star)
. More interestingly, as shown in Fig.~\ref{fig:subfig:6}, even when we apply stimulative training with 6 sub-networks (red five-pointed star) to MobileNetV3, the performance of all the sub-networks (red solid circle) can be also greatly improved compared to that of common training (blue solid circle). Besides, As shown in Fig.~\ref{fig:diff_training_strategy}, with more sub-networks involved in stimulative training, the network loafing problem could be relieved better, and the performance of a given residual network and all of its sub-networks could be improved more significantly. More similar results of different datasets and residual networks can be found in Appendix A. 
\subsection{Experiment 2: Collecting statistical characteristics of the performance of all residual sub-networks} \vspace{-1mm}
To demonstrate that stimulative training can improve the performance of a given residual network and all of its sub-networks, we collect various statistical characteristics of the performance of all residual sub-networks after the stimulative training and common training. As shown in Fig.~\ref{fig:statistical}, MobileNetV3 with common training (i.e., sample 0 sub-networks) has the lowest maximum value, mean value, and the highest standard deviation, which is mainly caused by the network loafing problem. As a contrast, MobileNetV3 with stimulative training (i.e., sample 144 sub-networks) has the highest maximum value, mean value, and the lowest standard deviation. Moreover, we can see that, MobileNetV3 using stimulative training with more sub-networks can usually provide higher maximum value, mean value and lower standard deviation. All these results indicate that stimulative training can improve the performance of all residual sub-networks effectively and efficiently. 

\subsection{Experiment 3: Destructing residual networks at test time} 
\begin{figure}[t]
\vspace{-4mm}
  \centering
  \subfigure[deleting one]{
    \label{fig:subfig:del_one} 
    \includegraphics[width=0.32\linewidth]{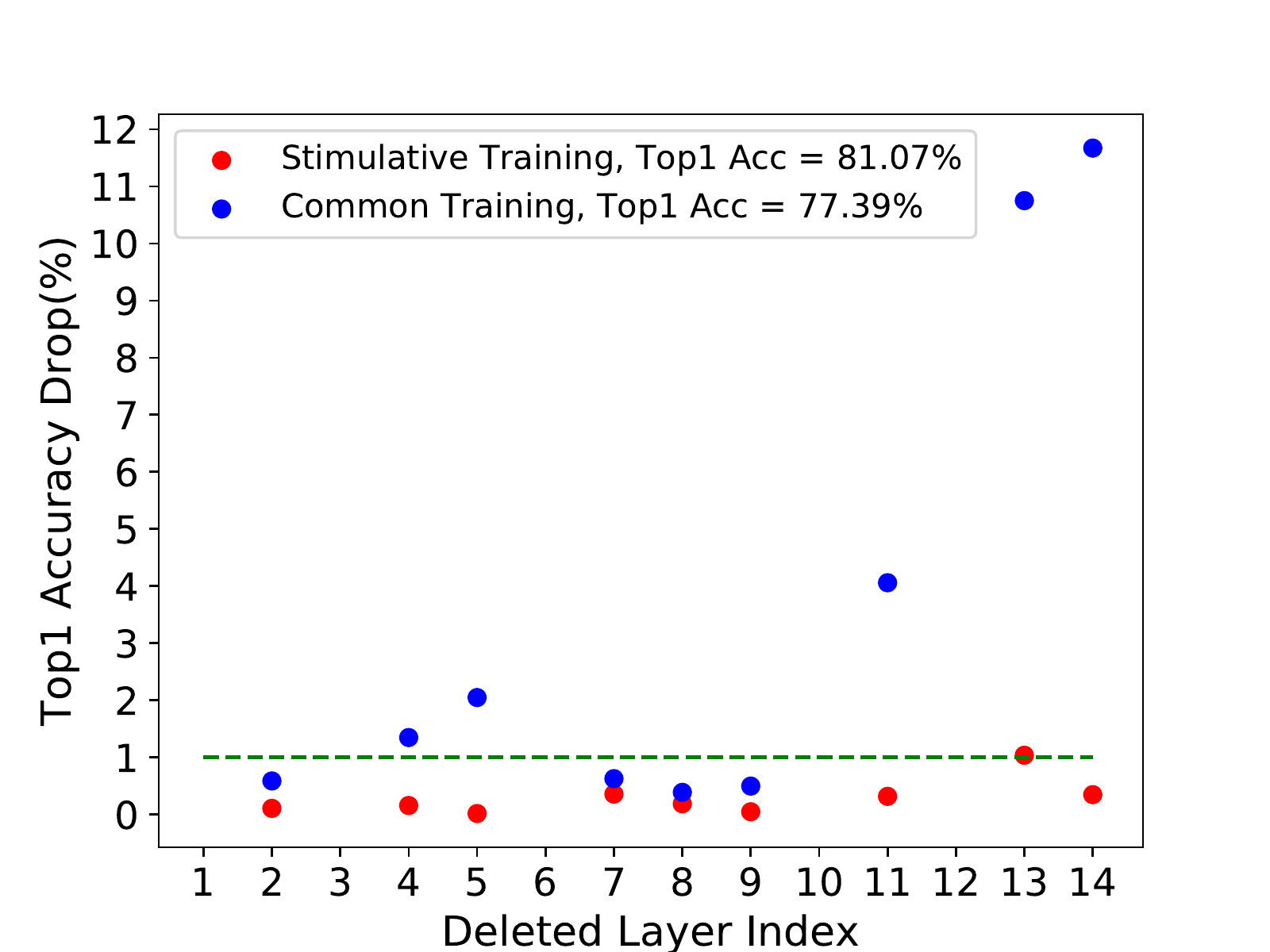}}
  \subfigure[deleting more]{
    \label{fig:subfig:del_more} 
    \includegraphics[width=0.32\linewidth]{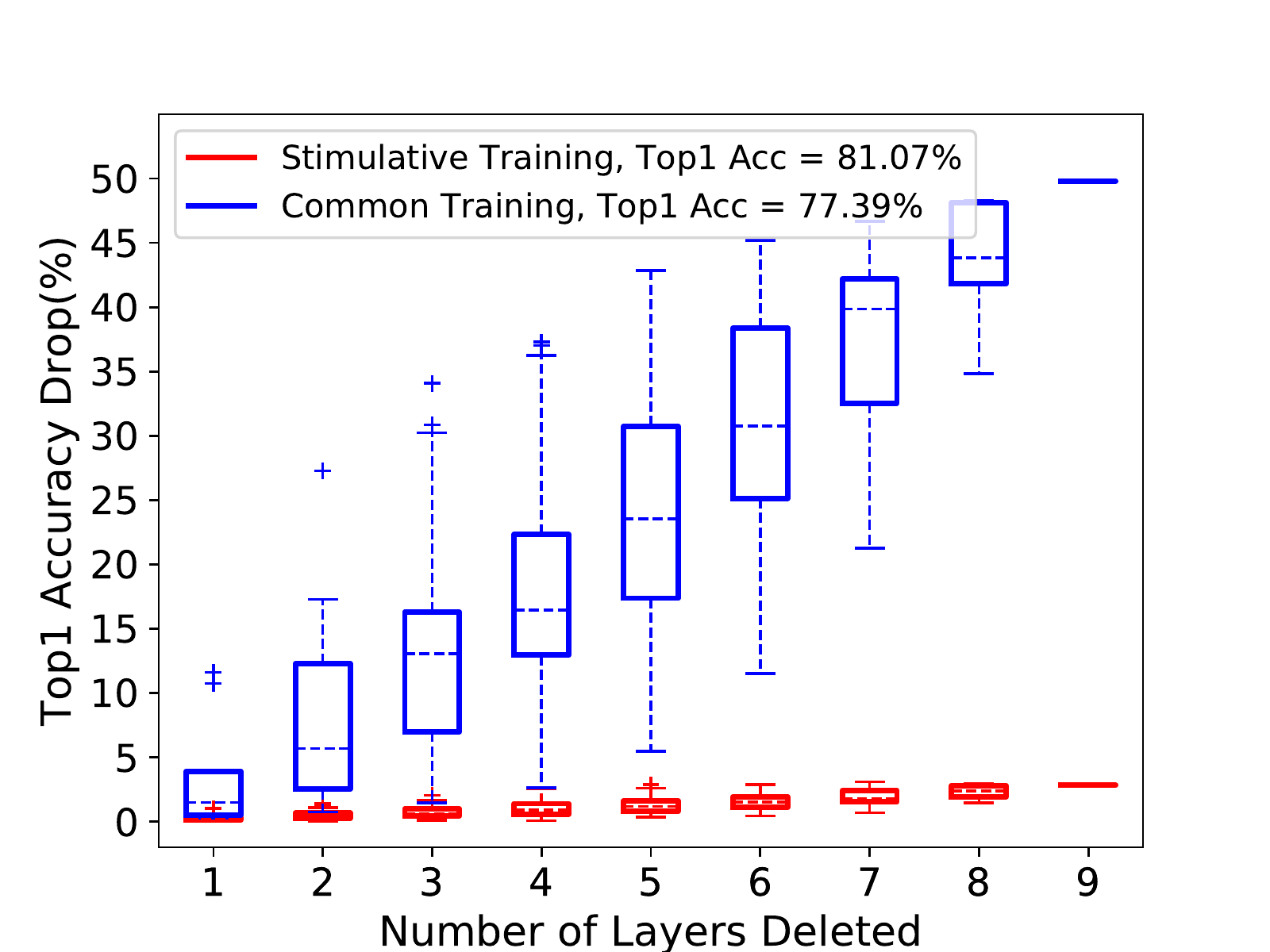}}
  \subfigure[permuting]{
    \label{fig:subfig:permuting} 
    \includegraphics[width=0.32\linewidth]{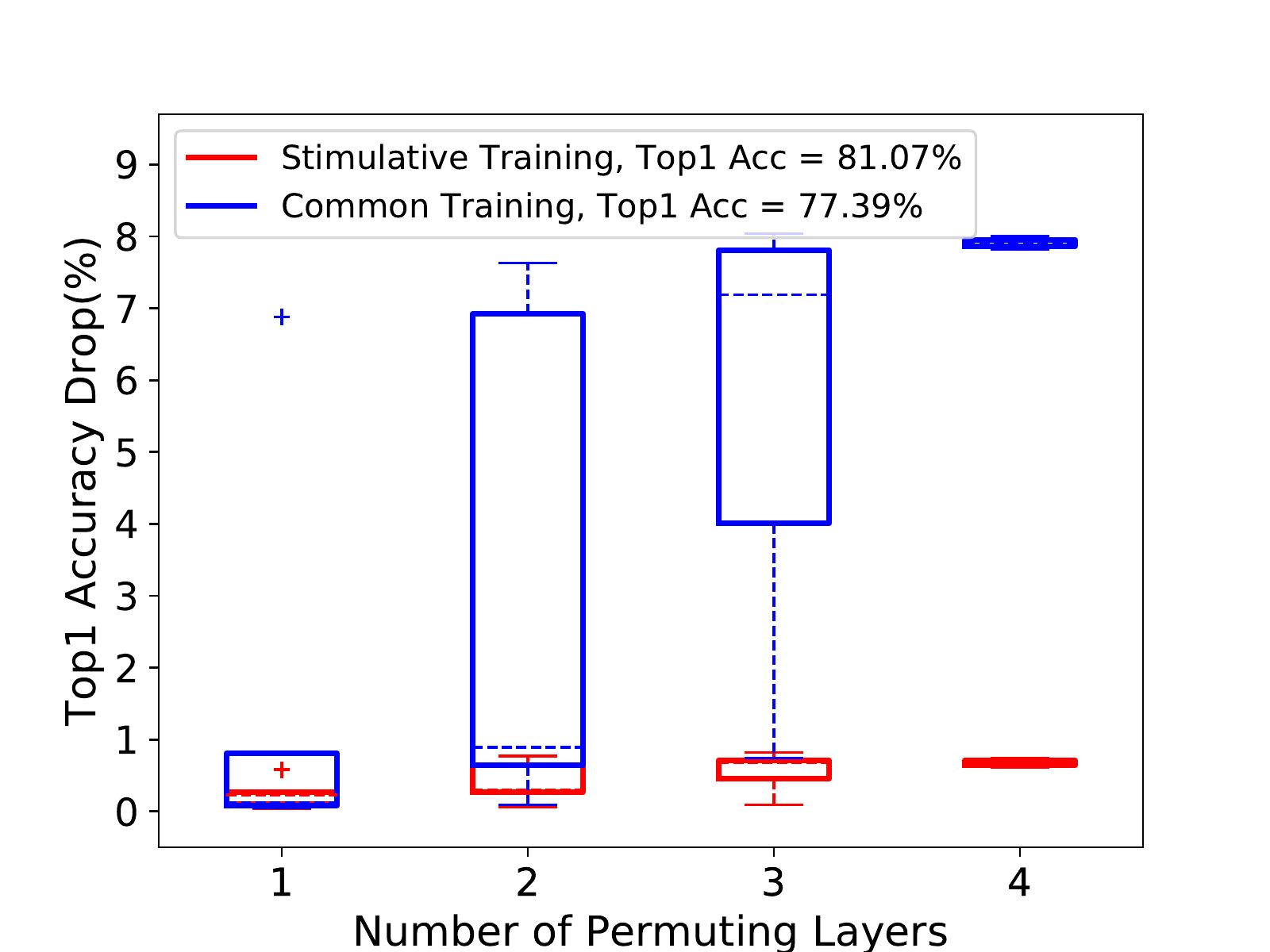}}
  \vspace{-3mm}
  \caption{Top1 accuracy drop when (a) deleting one layer, (b) deleting more layers and (c) permuting layers from MobileNetV3 with stimulative/common training on CIFAR100 dataset. Stimulative training can provide stronger robustness in resisting various network destruction operations than common training.}
  \label{fig:destruction} 
\vspace{-4mm}
\end{figure}

\vspace{-1mm}
In this experiment, we verify that stimulative training can provide stronger robustness in resisting various network destruction operations including deleting individual layers, deleting more layers and permuting layers than common training.
\paragraph{Deleting individual layers.} We first show the performance drop when deleting individual layers from trained MobileNetV3 at test time. As shown in Fig.~\ref{fig:subfig:del_one}, when applying common training to MobileNetV3, the performance drop is non-negligible when some layers are deleted (especially for the last several layers of each stage), and the trained MobileNetV3 is sensitive to the index of deleted layer. As a comparison, when applying stimulative training to MobileNetV3, no matter which layer is deleted, the performance drop of stimulative trained model is always lower than that of the common trained model, and the performance drop is always smaller than or equal to 1\%. Such results verify that stimulative training can reduce the dependence between sub-networks significantly.
\paragraph{Deleting more layers.} We then show the performance drop when deleting more layers from trained MobileNetV3 at test time. Note that, deleting more layers may come in a variety of combinations, thus we show the statistical box plot of performance drops for all combinations. As shown in Fig.~\ref{fig:subfig:del_more}, when applying common training to MobileNetV3, the medium value, minimum value and maximum value of performance drop for all combinations keeps continuously rising as the number of deleted layers increases. And the difference in performance drop for different combinations when removing the same number of layers can be very large. For example, when deleting 4 layers of common trained MobileNetV3, the Top1 accuracy drop for different combinations can vary from less than 5\% to higher than 35\%. Differently, when applying stimulative training, no matter how many layers are deleted, the values of performance drop for all combinations always keep very low (i.e., no more than 3\%), indicating that stimulative training can bring strong robustness in resisting network destruction.

\paragraph{Permuting layers.} We further show the performance drop when permuting layers in trained MobileNetV3 at test time. Note that, only the layers in the same stage can be permuted since the resolution and width of different stages are varying, and we show the detailed experimental settings in Appendix C.3. As shown in Fig.~\ref{fig:subfig:permuting}, when applying common training to MobileNetV3, the medium value, minimum value and maximum value of performance drop for all combinations increases with the rising number of permuted layers, and the variance of performance drop for different combinations when permuting the same number of layers may be also very large. As a comparison, the values of performance drop for all permuting combinations always keep very low (i.e., less than 1\%), when stimulative training is applied. This further verifies the superiority of stimulative training.

\subsection{Experiment 4: Recording the KL divergence between a residual network and all of its sub-networks when training} \vspace{-1mm}
\begin{minipage}[b]{0.5\linewidth}
To study the consistency between the output of a given residual network and all of its sub-networks, we record their KL distance in the whole training process. As shown in Fig.~\ref{fig:kl_distance}, when applying common training to MobileNetV3, the variance, maximum and medium value of KL distance between a given residual network and all of its sub-networks continuously increases with rising training epoch. For comparison, when stimulative training is applied to MobileNetV3, all values of KL distance between MobileNetV3 and all of its sub-networks keep very small (i.e., no more than 0.25) in the whole training process. Such results imply that the proposed stimulative training can maintain the consistency between the output of a given residual network and all of its sub-networks.
\end{minipage}
\hfill
\begin{minipage}[b]{0.45\linewidth}
\vspace{-6mm}
\begin{figure}[H]
\centering
\includegraphics[width=0.90\textwidth]{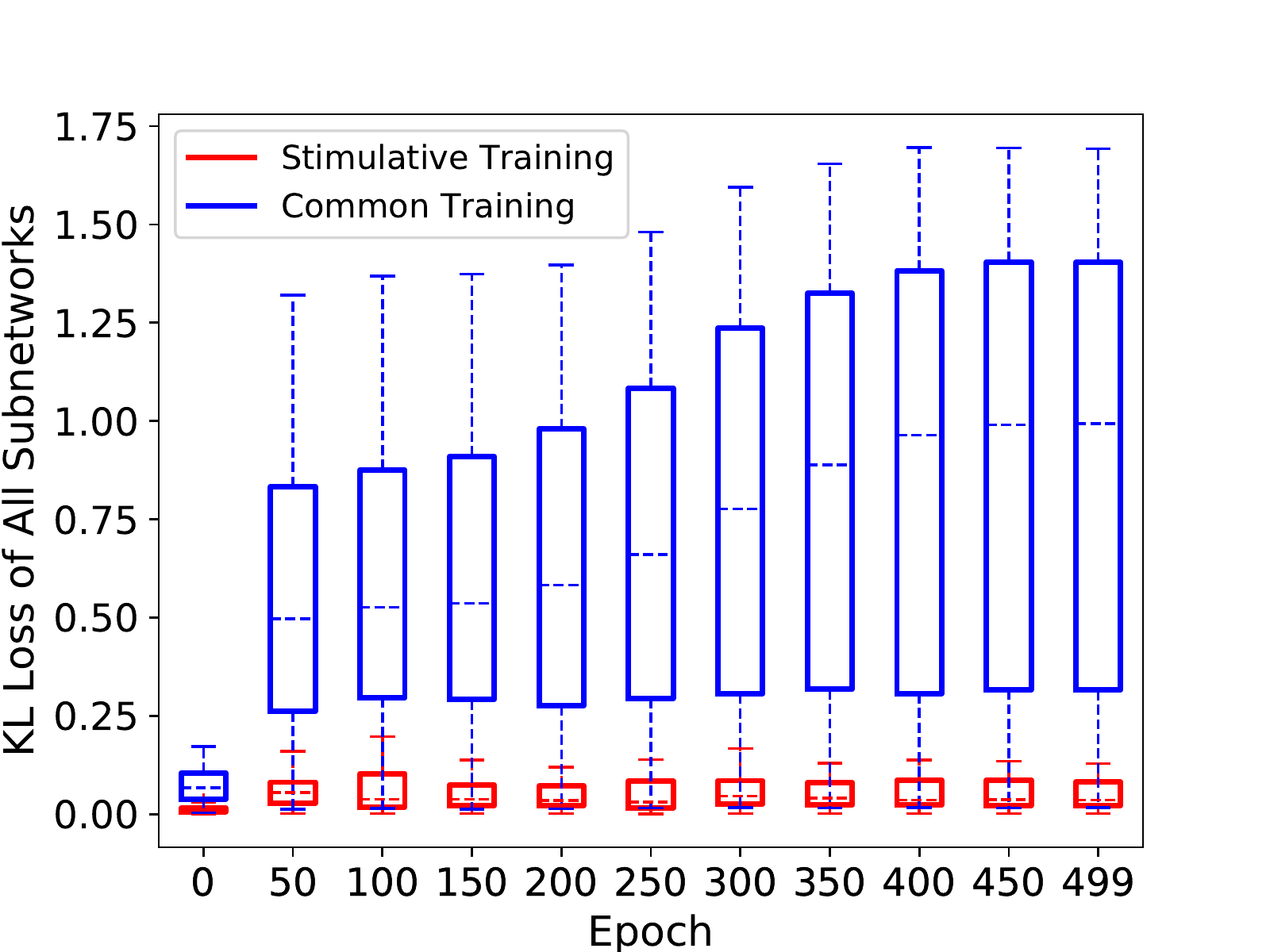}
\vspace{-3mm}
\caption{The distribution of KL distance between MobileNetV3 and all of its sub-networks in the whole training process. We train MobileNetV3 on CIFAR100 with stimulative and common schemes.}
\label{fig:kl_distance}
\end{figure}
\end{minipage}


\section{Theoretical Analysis}
\paragraph{Analysis 1: Why the performance of sub-networks can be improved.}
Based on Eq.~\ref{fomulation:ut loss}, our method can be treated as optimizing $CE(\mathcal{Z}(\theta_{D_m},x ),y)$ and $KL(\mathcal{Z}(\theta_{D_m},x),\mathcal{Z}(\theta_{D_s} ,x))$ at the same time. According to the convergence of stochastic gradient descent, after enough number of iterations, both losses can be bounded by a constant, which is formulated as
\begin{align}
&CE(\mathcal{Z}(\theta_{D_m},x ),y)=-\sum_{i=1}^{N} y_i \log p_{i}^m < \epsilon_1 \\
&KL(\mathcal{Z}(\theta_{D_m},x),\mathcal{Z}(\theta_{D_s} ,x))=\mathbb{E}_\Theta\left [  \sum_{i=1}^{N} p_{i}^m \log \frac{p_{i}^m}{p_{i}^s} \right ] < \epsilon_2  
\end{align}
where $N$ is the number of categories. The $\epsilon_1$ and $\epsilon_2$ are the tiny constants for corresponding losses. $\Theta$ is the set of all sub-networks. The $p^m$ and $p^s$ are the classification probability of main network and sub-network, respectively. Further, we can prove that \textit{the absolute difference of cross entropy of sub-networks and main network can be bounded by a constant} (detailed proof can be found in Appendix B), which can be formulated as
\begin{align}
\label{formulation:ce gap} \left |   CE(\mathcal{Z}(\theta_{D_m},x ),y)- \mathbb E _\Theta \left [  CE(\mathcal{Z}(\theta_{D_s},x ),y)\right ]\right | 
<\frac{\epsilon_2 + \log N }{e^{-\epsilon_1}} + \epsilon_1 
\end{align}
where $\frac{\epsilon_2 + \log N }{e^{-\epsilon_1}} + \epsilon_1$ is a small constant after optimization. Eq.~\ref{formulation:ce gap} demonstrates that the performance gap between sub-networks and the main network is constrained. Thus, \textit{as the rising of the main network performance, sub-networks performance will be improved step by step}. 
\paragraph{Analysis 2: Why the performance of given residual networks can be improved.} Neural network training is in fact a form of polynomial regression, and any neural network with activation functions can roughly correspond to a fitted polynomial~\cite{cheng2018polynomial}. For simplifying notations, suppose that both input $x$ and output $y$ in a neural network have only one element. Then, we can consider the training of a neural network as the fitting of a polynomial function.
\begin{align}
\label{fomulation:polynomial} 
y=F(x)=c_{0}+c_{1}x+c_{2}x^{2}+\cdots+c_{n} x^{n}+\ldots
\end{align}
where $c_{1}, \ c_{2},\ \ldots,\ c_{n},\ \ldots$ represent the polynomial coefficients. Based on training samples, we can establish the polynomial (neural network), which denotes the mapping relationship between inputs and outputs. For a test sample $x$, there always exists a point $x_{0}$ (e.g., a training sample) close enough to $x$, which can obtain an accurate output via the trained model. According to Taylor expansion, $F(x)$ at $x_{0}$ can be computed as :
\begin{align}
\label{fomulation:taylor} 
\begin{array}{l}
y=F(x)=\frac{F(x_{0})}{0 !}+\frac{F^{\prime}(x_{0})}{1 !}(x-x_{0})+ \\
\frac{F^{\prime \prime}(x_{0})}{2 !}(x-x_{0})^{2}+\cdots+\frac{F^{(n)}(x_{0})}{n !}(x-x_{0})^{n}+R_{n}(x)
\end{array}
\end{align}
where $R_{n}(x)$ denotes the higher degree infinitesimal of $(x-x_{0})^{n}$. According to Eq.~\ref{fomulation:polynomial} and Eq.~\ref{fomulation:taylor}, it is obvious that different terms of the fitted polynomial have different effects, and the low-degree terms tend to have greater effects on the predicted performance. Based on this point, ~\cite{sun2022low} further figures out that shallow sub-networks in residual networks roughly correspond to low-degree polynomials consisting of low-degree terms, while deep sub-networks are opposite. Following this theory, the proposed stimulative training can not only \textit{strengthen the learning of different items} of the polynomial (neural network), but also \textit{pay more attention to the learning of low-degree items}. As a result, the final performance of given residual networks can be improved by stimulative training.
 
\section{Verification on Various Datasets and Networks}
\paragraph{CIFAR implementation details.}
CIFAR\cite{krizhevsky2009learning} is a classical image classification dataset consisting of 50,000 training images and 10,000 testing images. It includes CIFAR-100 in 100 categories and CIFAR-10 in 10 categories. For MobileNetV3 and ResNet50, the data augmentations follow \cite{sahni2021compofa}, we use SGD optimizer and train the model for 500 epochs with a batch size of 64. The initial learning rate is 0.05 with cosine decay schedule. The weight decay is $3 \times 10^{-5}$ and momentum is 0.9.
\paragraph{ImageNet implementation details.}
We implement our method on large-scale ImageNet\cite{deng2009imagenet} dataset containing 1.2 million training images and 50,000 validation images from 1,000 categories. For ResNet families, we use standard data augmentations for ImageNet, as done in \cite{szegedy2015going,huang2017densely,ghiasi2018dropblock}. We utilize SGD optimizer to train the model for 100 epochs with a batch size of 512, and the learning rate is 0.2 with cosine decay schedule. The weight decay is $1 \times 10^{-4}$ and momentum is 0.9.

\begin{table}[]
\vspace{-2mm}
\caption{The main network and all sub-networks performance (\%) of stimulative training (ST) compared with common training (CT), on various residual networks and datasets.}
\vspace{-2mm}
\centering
\label{table:performance of network}
\begin{tabular}{|l|l|l|l|l|l|}
\hline
method   & MBV3\_C10    & MBV3\_C100   & Res50\_C100  & Res50\_Img   & Res101\_Img  \\ \hline
CT(main) & 95.72        & 77.39        & 76.53        & 76.1         & 77.37        \\ \hline
ST(main) & \textbf{96.88(+1.16)} & \textbf{81.07(+3.68)} & \textbf{81.06(+4.43)} & \textbf{77.22(+1.12)} & \textbf{78.58(+1.21)} \\ \hline
CT(all)  & 78.13±14.29  & 55.26±13.37  & 34.03±18.92  & 39.46±16.82  & 35.78±18.84  \\ \hline
ST(all)  & \textbf{96.21±0.43}   & \textbf{80.01±0.59}   & \textbf{79.66±1.77}   & \textbf{73.3±2.89}    & \textbf{75.52±3.6}    \\ \hline
\end{tabular}
\vspace{-4mm}
\end{table}

\paragraph{Main network and sub-networks performance.}
We further verify the effectiveness of stimulative training (ST) on various networks and datasets.
Table.\ref{table:performance of network} shows its comparison with common training (CT).
Besides the Top1 accuracy of main network, we also report the the mean value and standard deviation of Top1 accuracy for all sub-networks.
It is obvious that stimulative training can noticeably improve the performance of various networks on various datasets compared to common training (e.g., by 4.43\% for ResNet50 on CIFAR-100 and 1.21\% for ResNet101 on ImageNet). Moreover, the mean performance of all sub-networks is dramatically improved and standard deviation greatly drops. For example, 
stimulative training for MobileNetV3 on CIFAR100 has a mean performance of 80.01\% (much higher than common training of 55.26\%) and a standard deviation of 0.59\% (much lower than 13.37\%). Such results validate that stimulative training can well alleviate the network loafing problem, and improve the performance of the given residual network and all of its sub-networks.
Experimental results on various models and datasets verify the generalization of stimulative training, we advocate utilizing it as a general technology to train residual networks.                         

\section{Conclusions}
The paper understands and improves residual networks from a social psychology perspective of loafing. Based on the novel view that a residual network behaves like an ensemble network, we find that various residual networks invariably exhibit loafing-like behaviors that are consistent with the social loafing problem of social psychology. We define this previously overlooked problem as network loafing.
As the loafing problem hinders the productivity of each individual and the whole collective, we learn from social psychology and propose a stimulative training strategy to solve the network loafing problem. Comprehensive empirical analyses show that stimulative training can improve the performance of a given residual network and all of its sub-networks, and provide strong robustness in resisting various network destruction operations. Furthermore, we theoretically show why the performance of a given residual network and its sub-networks can be improved. Experiments on various datasets and residual networks demonstrate the effectiveness of the proposed method.

\section{Limitations}
The proposed method suffers from about 1.4 times of computation cost of the original model training to get better performance and robustness. As the first research of network loafing problem, the proposed method is a positive pioneer-like exploration. We believe designing a more efficient method to solve the network loafing problem is a worthy research direction in the future.

Residual structure is widely applied in numerous different types of models including DenseNet and transformer. It will be of vital value to study whether the loafing problem exists in these models and explore the proper method to solve this problem. Since most of existing works overlook the loafing problem, it may also be a feasible way to apply the proposed method in pretrained models. What’s more, the proposed method adopts main network logits as supervision to alleviate loafing and analogously, the proposed method may be applied in self-supervised learning.

We believe that taking full advantage of splendid achievements in interdisciplinary research can help promote the development of deep learning. We hope that this paper can provide a new perspective to inspire more researchers to comprehend and improve deep neural networks from other fields such as social psychology.

\section{Acknowledgements}
This work is supported by National Natural Science Foundation of China (No. U1909207 and 62071127), and Shanghai Municipal Science and Technology Major Project (No.2021SHZDZX0103). Wanli Ouyang is supported by the Australian Research Council Grant DP200103223, Australian Medical Research Future Fund MRFAI000085, CRC-P Smart Material Recovery Facility (SMRF) – Curby Soft Plastics, and CRC-P ARIA - bionic visual-spatial prosthesis for the blind.

\bibliography{neurips_2022}






\section*{Checklist}

The checklist follows the references.  Please
read the checklist guidelines carefully for information on how to answer these
questions.  For each question, change the default \answerTODO{} to \answerYes{},
\answerNo{}, or \answerNA{}.  You are strongly encouraged to include a {\bf
justification to your answer}, either by referencing the appropriate section of
your paper or providing a brief inline description.  For example:
\begin{itemize}
  \item Did you include the license to the code and datasets? \answerYes{See Section~\ref{gen_inst}.}
  \item Did you include the license to the code and datasets? \answerNo{The code and the data are proprietary.}
  \item Did you include the license to the code and datasets? \answerNA{}
\end{itemize}
Please do not modify the questions and only use the provided macros for your
answers.  Note that the Checklist section does not count towards the page
limit.  In your paper, please delete this instructions block and only keep the
Checklist section heading above along with the questions/answers below.

\begin{enumerate}

\item For all authors...
\begin{enumerate}
  \item Do the main claims made in the abstract and introduction accurately reflect the paper's contributions and scope?
    \answerYes{}
  \item Did you describe the limitations of your work?
    \answerNo{}
  \item Did you discuss any potential negative societal impacts of your work?
    \answerNo{}
  \item Have you read the ethics review guidelines and ensured that your paper conforms to them?
    \answerYes{}
\end{enumerate}

\item If you are including theoretical results...
\begin{enumerate}
  \item Did you state the full set of assumptions of all theoretical results?
    \answerYes{}
        \item Did you include complete proofs of all theoretical results?
    \answerYes{}
\end{enumerate}

\item If you ran experiments...
\begin{enumerate}
  \item Did you include the code, data, and instructions needed to reproduce the main experimental results (either in the supplemental material or as a URL)?
    \answerYes{}
  \item Did you specify all the training details (e.g., data splits, hyperparameters, how they were chosen)?
    \answerYes{}
        \item Did you report error bars (e.g., with respect to the random seed after running experiments multiple times)?
    \answerNo{}
        \item Did you include the total amount of compute and the type of resources used (e.g., type of GPUs, internal cluster, or cloud provider)?
    \answerNo{}
\end{enumerate}

\item If you are using existing assets (e.g., code, data, models) or curating/releasing new assets...
\begin{enumerate}
  \item If your work uses existing assets, did you cite the creators?
    \answerYes{}
  \item Did you mention the license of the assets?
    \answerYes{}
  \item Did you include any new assets either in the supplemental material or as a URL?
    \answerYes{}
  \item Did you discuss whether and how consent was obtained from people whose data you're using/curating?
    \answerNA{}
  \item Did you discuss whether the data you are using/curating contains personally identifiable information or offensive content?
    \answerNA{}
\end{enumerate}

\item If you used crowdsourcing or conducted research with human subjects...
\begin{enumerate}
  \item Did you include the full text of instructions given to participants and screenshots, if applicable?
    \answerNA{}
  \item Did you describe any potential participant risks, with links to Institutional Review Board (IRB) approvals, if applicable?
    \answerNA{}
  \item Did you include the estimated hourly wage paid to participants and the total amount spent on participant compensation?
    \answerNA{}
\end{enumerate}

\end{enumerate}


\appendix


\end{document}


\bibliographystyle{unsrt} 

\maketitle

\appendix

\section*{Appendix A: The loafing problem of different datasets and residual networks}
\begin{figure}[H]
\vspace{-4mm}
  \centering
  \subfigure[MBV3 on C10]{
    \label{fig:subfig:loafing_mbv3_c10} 
    \includegraphics[width=0.31\linewidth]{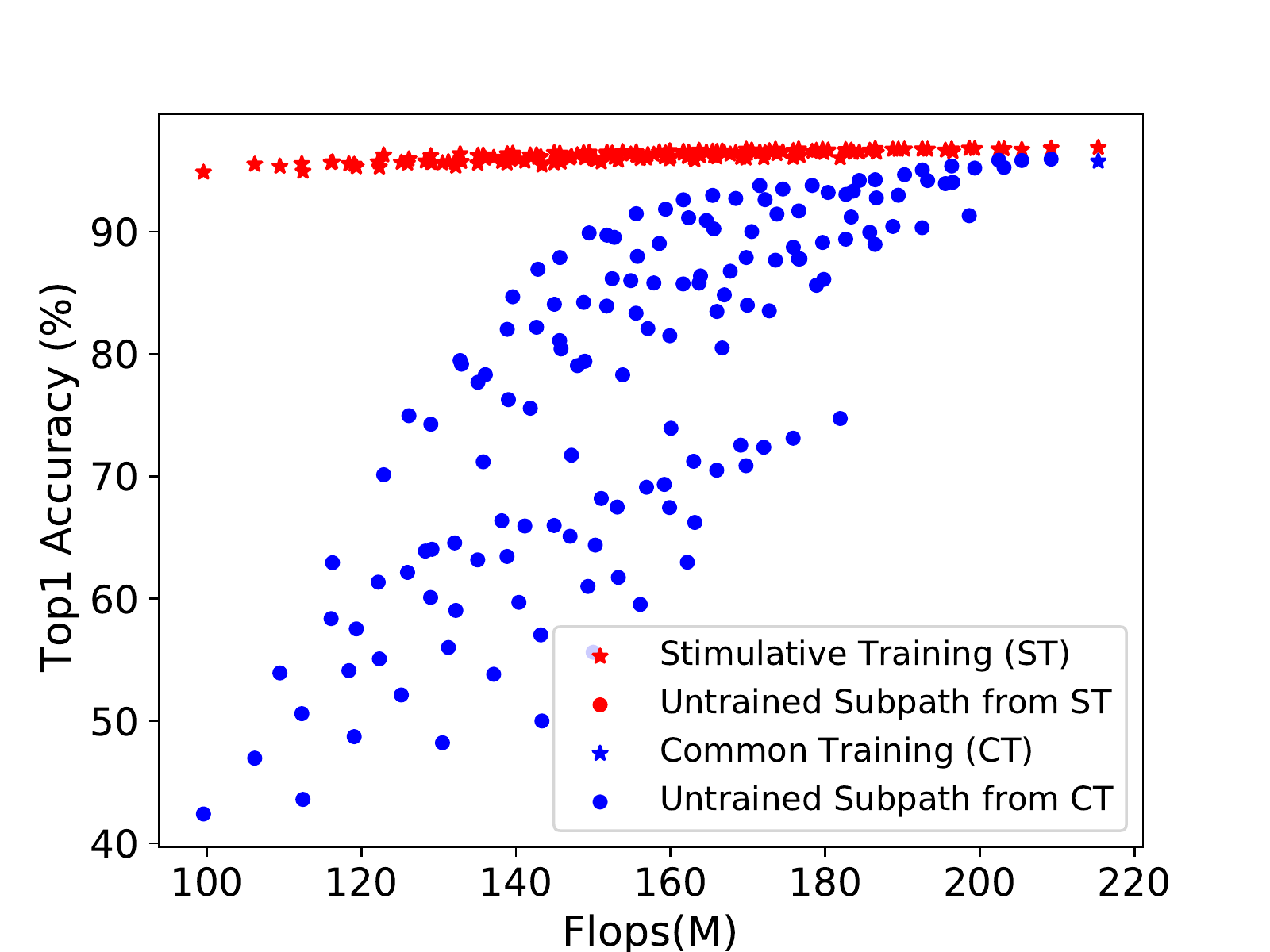}}
  \subfigure[Res50 on C100]{
    \label{fig:subfig:loafing_res50_c100} 
    \includegraphics[width=0.31\linewidth]{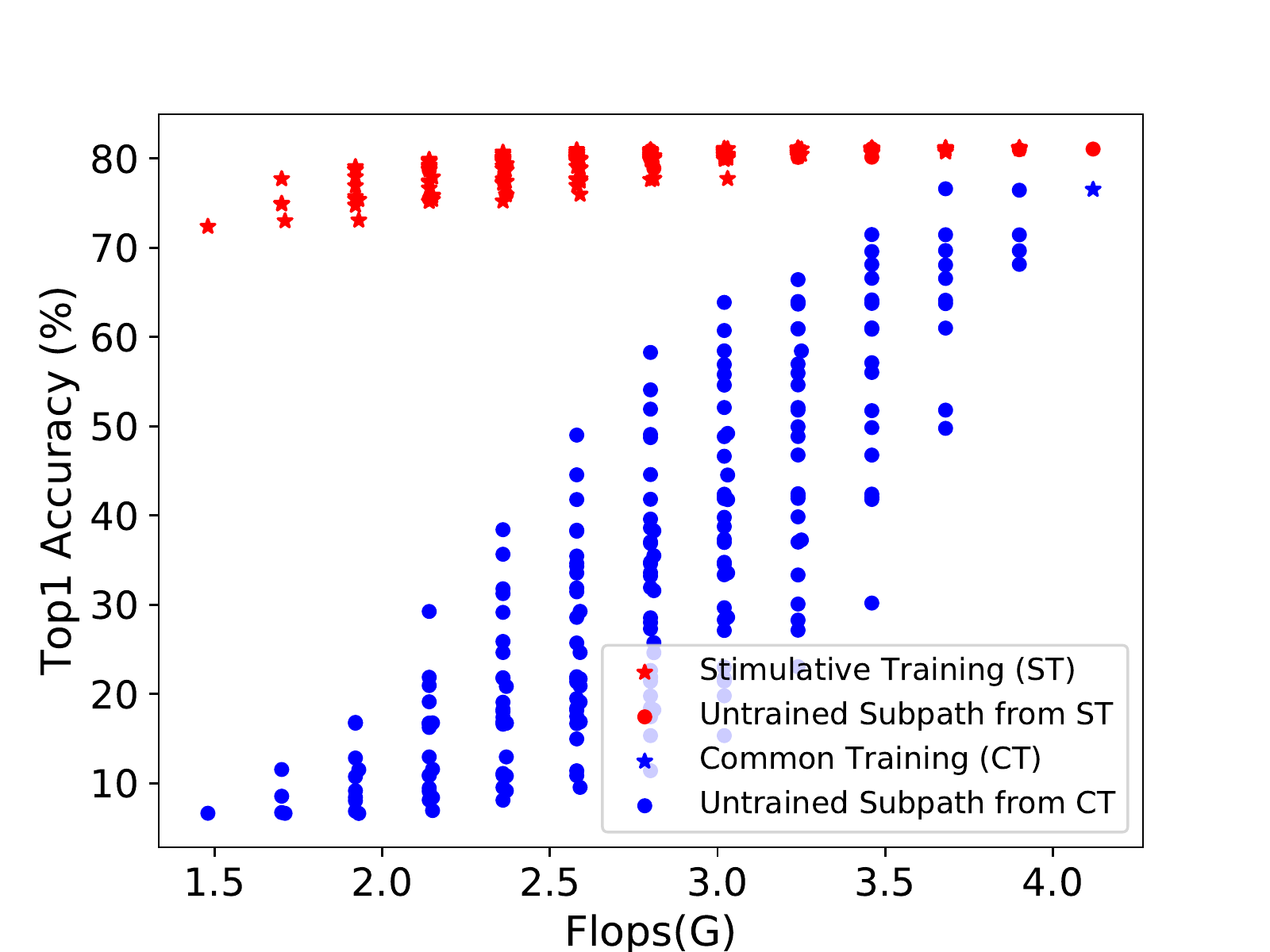}}
  \subfigure[Res50 on ImageNet]{
    \label{fig:subfig:loafing_res50_img} 
    \includegraphics[width=0.31\linewidth]{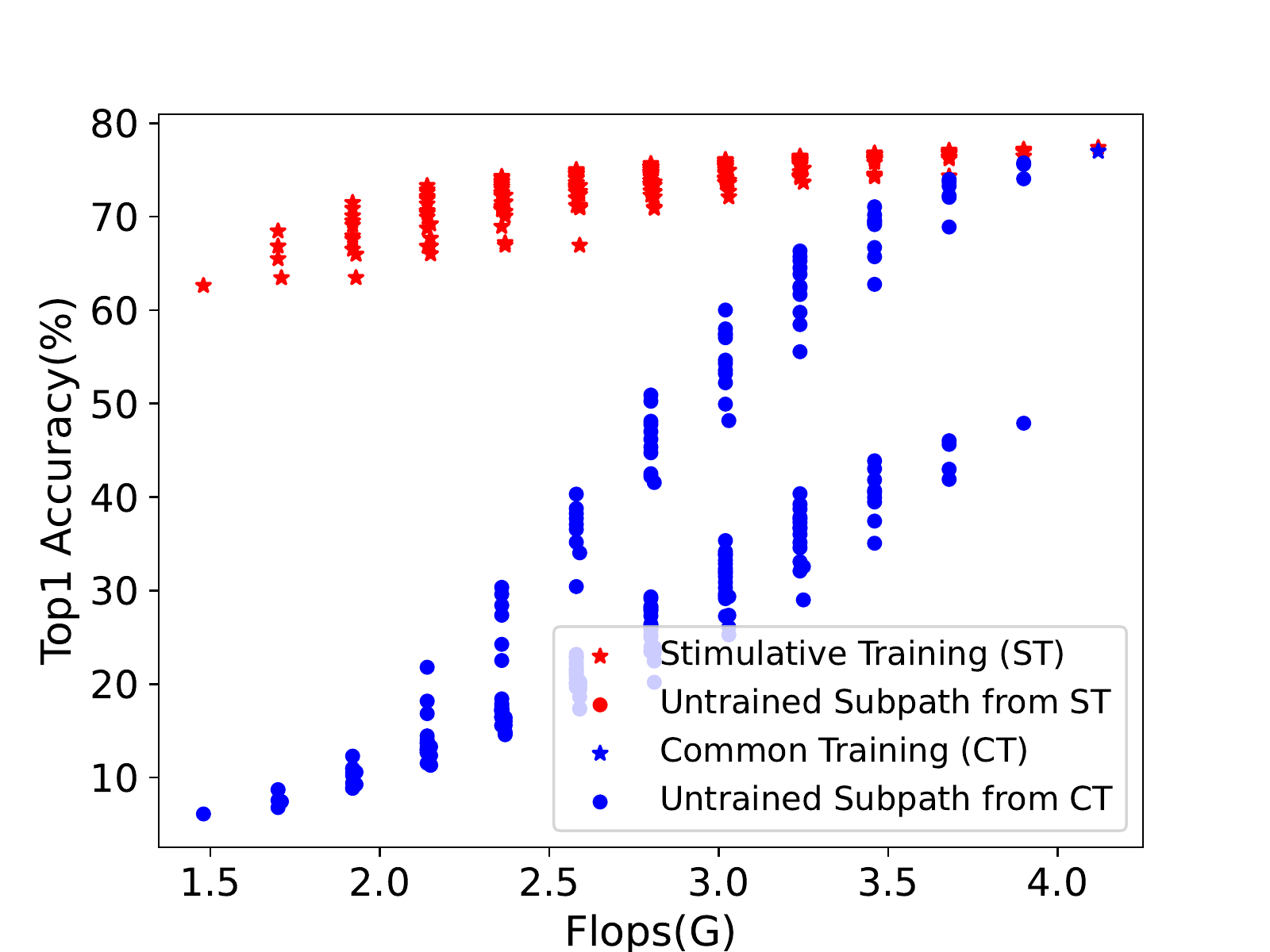}}
  \vspace{-4mm}
  \caption{Stimulative training can improve the performance of a given residual network and all of its sub-networks significantly. We further verify it on various residual networks and benchmark datasets.}
  \label{fig:loafing_diff}
\vspace{-4mm}
\end{figure}

We further verify that stimulative training can well handle the loafing problem on different datasets and residual networks. As shown in Fig.~\ref{fig:loafing_diff}, we can see that stimulative training can always improve the performance of a given residual network and all of its sub-networks by a larger margin on various residual networks and benchmark datasets. In other words, different residual networks trained on different datasets invariably suffer from the problem of network loafing, which can be well solved by the proposed stimulative training strategy.

\section*{Appendix B: Proof of theoretical analysis 1}
According to the convergence of SGD, cross entropy loss $CE(\mathcal{Z}(\theta_{D_m},x ),y)$ and KL-divergence loss $KL(\mathcal{Z}(\theta_{D_m},x),\mathcal{Z}(\theta_{D_s} ,x))$ used in our method can be bounded by a tiny constant, expressed as 
\begin{align}
&\label{formulate:ce loss} CE(\mathcal{Z}(\theta_{D_m},x ),y)=-\sum_{i=1}^{N} y_i \log p_{i}^m < \epsilon_1, \\
&\label{formulate:kl loss}KL(\mathcal{Z}(\theta_{D_m},x),\mathcal{Z}(\theta_{D_s} ,x))=\mathbb{E}_\Theta\left [  \sum_{i=1}^{N} p_{i}^m \log \frac{p_{i}^m}{p_{i}^s} \right ] < \epsilon_2, \\ 
& 0<p_{i}^m,p_{i}^s<1, \sum_{i=1}^{N}p_{i}^m=1, \sum_{i=1}^{N}p_{i}^s=1,
\end{align}
where $\theta_{D_m}$ and $\theta_{D_s}$ denote the learned weights of the main network and sub-network, respectively. $\mathcal{Z}$ is the network output, $x$ is the input image, and $y$ is the label. $p_{i}^m$ and $p_{i}^s$ are the prediction probability of the main network and sub-network, respectively.  We define $k$ as the ground truth index (i.e., $y_k=1$ and $y_i=0 (i \ne k) $), $\Theta$ as the set of sampled sub-networks, $\epsilon_1$ and $\epsilon_2$ as the tiny constants. Therefore, our target is to prove that the gap between the $CE$ loss of the main network and all sub-networks is bounded by a tiny constant, written as
\begin{align}
& \label{formulation:ce gap} \left |   CE(\mathcal{Z}(\theta_{D_m},x ),y)- \mathbb E _\Theta \left [  CE(\mathcal{Z}(\theta_{D_s},x ),y)\right ]\right | 
\\
& \label{formulation:target}= \left | \mathbb E _\Theta \left [\sum_{i=1}^{N}y_i \log \frac{p_i^m}{p_i^s}   \right ] \right | = \left | \mathbb E _\Theta \left [ \log \frac{p_k^m}{p_k^s} \right ] \right |<\epsilon_3.
\end{align}
Now, we begin to prove the existence of $\epsilon_3$. For Formulation.~\ref{formulate:ce loss}, it can be simplified to $-\log p_k^m \le \epsilon_1$, which can be equally written as 
\begin{align}
\label{formulation:key1}p_k^m \ge  e^{-\epsilon_1}.
\end{align}
For Formulation.~\ref{formulate:kl loss}, it can be equally written as 
\begin{align}
&-H(p^m)-\mathbb{E}_\Theta\left [  \sum_{i=1}^{N} p_{i}^m \log p_{i}^s \right ] < \epsilon_2,\\
&\label{formulation:key2}-\mathbb{E}_\Theta\left [  \sum_{i=1}^{N} p_{i}^m \log p_{i}^s \right ] < \epsilon_2+H(p^m)<\epsilon_2+\log N,
\end{align}
where $H(\cdot)$ represents the entropy operator. Further, based on Formulation.~\ref{formulation:key1} and Formulation.~\ref{formulation:key2}, Formulation.~\ref{formulation:target} can be bounded by 
\begin{align}
&\label{fomulation:proof1}\left | \mathbb E _\Theta \left [ \log \frac{p_k^m}{p_k^s} \right ] \right |=\left | \log p_k^m - \mathbb E _\Theta \left [ \log {p_k^s} \right ] \right |
<\left | \log p_k^m \right |+\left | E _\Theta \left [ \log {p_k^s} \right ]\right |\\
&\label{fomulation:proof2}<\epsilon_1 + \left | \frac{E _\Theta \left [ p_k^m\log p_k^s \right ]}{p_k^m} \right |<\epsilon_1 + \left | \frac{E _\Theta \left [ \sum_{i=1}^{N} p_i^m\log p_i^s \right ]}{p_k^m} \right |\\
&\label{fomulation:proof3}<\epsilon_1 + \left | \frac{\epsilon_2+\log N }{p_k^m} \right |<\epsilon_1 + \frac{\epsilon_2+\log N }{e^{-\epsilon _1}}.
\end{align}
In Formulation. \ref{fomulation:proof1}, we utilize triangle inequality. In Formulation. \ref{fomulation:proof2} and \ref{fomulation:proof3}, we scale the inequality according to Formulation. \ref{formulation:key2} and \ref{formulation:key1}. As $\epsilon_3 = \epsilon_1 + \frac{\epsilon_2+\log N }{e^{-\epsilon _1}}$ is a tiny constant which is independent of $p$, we finish the proof.

\section*{Appendix C: More details about experimental settings}
\subsection*{Appendix C.1: The procedure of stimulative training}
We show the procedure of stimulative training in Alg.~\ref{ code:stimulative training }. In short, we randomly sample an ordered residual sub-network in the main network in each minibatch, and adopt KL divergence loss to constrain the output of the sub-network not far from that of the main network. Similar to solutions for preventing social loafing in social psychology, sampling ordered residual sub-networks aims to increase individual supervision sufficiently, and adopting KL divergence loss aims to make the goals of residual sub-networks and the given residual network more consistent. We will release our codes upon acceptance of this paper. 

\vspace{-5pt} 
\begin{algorithm}[H]
\caption{Stimulative Training}   
\label{alg:stimulative training}   
\begin{algorithmic}[1] 
\REQUIRE ~~\\ 
Main network $D_m$; Total training iterations $N$; Loss balanced coefficient $\lambda$; \\ Input $x$ and label $y$ of each minibatch; Random sampling $\pi$.
\STATE Construct the main network and initialize the main network weights $\theta_{D_m}$.
\label{ code:stimulative training }
\STATE For each $t\in \left[ 1, N \right]$ do   
\STATE ~~~Sample an ordered residual sub-network $D_s = \pi(D_m)$
\STATE ~~~Main network forwards $\mathcal Z _m=D_m(x,\theta_{D_m})$, and compute the loss \\ ~~~$\mathcal{L}_{m} = CE(\mathcal Z_m,y)$ 
\STATE ~~~Sub-network forwards $\mathcal Z _s=D_s(x,\theta_{D_s})$, and compute the loss \\ ~~~ $\mathcal{L}_{s} = KL(\mathcal Z _m,\mathcal Z _s)$
\STATE ~~~Compute the stimulative training loss, $\mathcal{L}_{st} =  \mathcal{L}_m + \lambda \mathcal{L}_s$
\STATE ~~~Backward and update network weights $\theta_{D_m}$ by descending $\nabla_{\theta_{D_m}} \mathcal{L}_{st}$
\STATE End.
\end{algorithmic} 
\end{algorithm} 
\vspace{-4pt} 

\subsection*{Appendix C.2: Settings of empirical analysis}
For empirical analysis, we train NAS-searched model MobileNetV3~\cite{howard2019searching} on CIFAR100 with common and stimulative training strategy, respectively, and test the diverse empirical characteristics of trained MobileNetV3 and all of its ordered residual sub-networks. For training, we use SGD optimizer and train the model for 500 epochs with a batch size of 64, the initial learning rate is 0.05 with cosine decay schedule, the weight decay is $3 \times 10^{-5}$ and momentum is 0.9. For testing, we always employ batchnorm re-calibration for each sampled sub-network following~\cite{yu2019universally}.

\subsection*{Appendix C.3: Settings of destructing residual network}
For the destruction experiment of residual networks, we employ the main body of MobileNetV3 for testing and ignore basic input/output layers, as shown in Table~\ref{tab:main body}. For deleting one or more layers, we may delete every layer except the downsampling layer considering that downsampling layers in MobileNetV3 are single branch structure. For permuting layers, we only permute the layers in the same stage since the resolution and width of different stages in MobileNetV3 are varying. We show the performance drop when conducting different number of permuting operations, and consider all combinations when conducting a fixed number of permuting operations. All possible combinations when conducting different number of permuting operations are shown in Table~\ref{tab:all combinations}.

\begin{table}[H]
\vspace{-1mm}
\caption{The main body of MobileNetV3. Input denotes the input size of feature maps. \#out is the output channel number and s is the stride. Index denotes the layer index when applying destruction.}
\label{tab:main body} 
\vspace{-1mm}
\centering
\setlength\tabcolsep{2pt}
\resizebox{1.0\columnwidth}{!}{
\begin{tabular}{c|cc|ccc|cccc|cc|ccc}
\hline
Input          & 112²×16    & 56²×24     & 56²×24     & 28²×40     & 28²×40     & 28²×40     & 14²×80     & 14²×80     & 14²×80     & 14²×80      & 14²×112     & 14²×112     & 7²×160      & 7²×160      \\
\#out          & 24         & 24         & 40         & 40         & 40         & 80         & 80         & 80         & 80         & 112         & 112         & 160         & 160         & 160         \\
s              & 2          & 1          & 2          & 1          & 1          & 2          & 1          & 1          & 1          & 1           & 1           & 2           & 1           & 1           \\ \hline
\textbf{Index} & \textbf{1} & \textbf{2} & \textbf{3} & \textbf{4} & \textbf{5} & \textbf{6} & \textbf{7} & \textbf{8} & \textbf{9} & \textbf{10} & \textbf{11} & \textbf{12} & \textbf{13} & \textbf{14} \\ \hline
\end{tabular}
}
\end{table}

\begin{table}[H]
\vspace{-3mm}
\caption{All combinations when conducting different number of permuting operations.}
\label{tab:all combinations} 
\vspace{-1mm}
\centering
\begin{tabular}{c|c}
\hline
Number of permuting layers     & All combinations when permuting layers                                                                                                                                                                                                                                                                                                                                                                                                                                                                                                                                                                                                                                                                                                                                                                                                                       \\ \hline
Number of permuting = 1 & \begin{tabular}[c]{@{}c@{}}{[}{[}{[}1, 2{]}, {[}3, 5, 4{]}, {[}6, 7, 8, 9{]}, {[}10, 11{]}, {[}12, 13, 14{]}{]},\\ {[}{[}1, 2{]}, {[}3, 4, 5{]}, {[}6, 8, 7, 9{]}, {[}10, 11{]}, {[}12, 13, 14{]}{]},\\ {[}{[}1, 2{]}, {[}3, 4, 5{]}, {[}6, 7, 9, 8{]}, {[}10, 11{]}, {[}12, 13, 14{]}{]},\\ {[}{[}1, 2{]}, {[}3, 4, 5{]}, {[}6, 9, 8, 7{]}, {[}10, 11{]}, {[}12, 13, 14{]}{]},\\ {[}{[}1, 2{]}, {[}3, 4, 5{]}, {[}6, 7, 8, 9{]}, {[}10, 11{]}, {[}12, 14, 13{]}{]}{]}\end{tabular}                                                                                                                                                                                                                                                                                                                                                     \\ \hline
Number of permuting = 2 & \begin{tabular}[c]{@{}c@{}}{[}{[}{[}1, 2{]}, {[}3, 5, 4{]}, {[}6, 8, 7, 9{]}, {[}10, 11{]}, {[}12, 13, 14{]}{]},\\ {[}{[}1, 2{]}, {[}3, 5, 4{]}, {[}6, 7, 9, 8{]}, {[}10, 11{]}, {[}12, 13, 14{]}{]},\\ {[}{[}1, 2{]}, {[}3, 5, 4{]}, {[}6, 9, 8, 7{]}, {[}10, 11{]}, {[}12, 13, 14{]}{]},\\ {[}{[}1, 2{]}, {[}3, 4, 5{]}, {[}6, 8, 7, 9{]}, {[}10, 11{]}, {[}12, 14, 13{]}{]},\\ {[}{[}1, 2{]}, {[}3, 4, 5{]}, {[}6, 7, 9, 8{]}, {[}10, 11{]}, {[}12, 14, 13{]}{]},\\ {[}{[}1, 2{]}, {[}3, 4, 5{]}, {[}6, 9, 8, 7{]}, {[}10, 11{]}, {[}12, 14, 13{]}{]},\\ {[}{[}1, 2{]}, {[}3, 4, 5{]}, {[}6, 8, 9, 7{]}, {[}10, 11{]}, {[}12, 13, 14{]}{]},\\ {[}{[}1, 2{]}, {[}3, 4, 5{]}, {[}6, 9, 7, 8{]}, {[}10, 11{]}, {[}12, 13, 14{]}{]},\\ {[}{[}1, 2{]}, {[}3, 5, 4{]}, {[}6, 7, 8, 9{]}, {[}10, 11{]}, {[}12, 14, 13{]}{]}{]}\end{tabular} \\ \hline
Number of permuting = 3 & \begin{tabular}[c]{@{}c@{}}{[}{[}{[}1, 2{]}, {[}3, 5, 4{]}, {[}6, 8, 7, 9{]}, {[}10, 11{]}, {[}12, 14, 13{]}{]},\\ {[}{[}1, 2{]}, {[}3, 5, 4{]}, {[}6, 7, 9, 8{]}, {[}10, 11{]}, {[}12, 14, 13{]}{]},\\ {[}{[}1, 2{]}, {[}3, 5, 4{]}, {[}6, 9, 8, 7{]}, {[}10, 11{]}, {[}12, 14, 13{]}{]},\\ {[}{[}1, 2{]}, {[}3, 5, 4{]}, {[}6, 8, 9, 7{]}, {[}10, 11{]}, {[}12, 13, 14{]}{]},\\ {[}{[}1, 2{]}, {[}3, 5, 4{]}, {[}6, 9, 7, 8{]}, {[}10, 11{]}, {[}12, 13, 14{]}{]},\\ {[}{[}1, 2{]}, {[}3, 4, 5{]}, {[}6, 8, 9, 7{]}, {[}10, 11{]}, {[}12, 14, 13{]}{]},\\ {[}{[}1, 2{]}, {[}3, 4, 5{]}, {[}6, 9, 7, 8{]}, {[}10, 11{]}, {[}12, 14, 13{]}{]}{]}\end{tabular}                                                                                                                                                                           \\ \hline
Number of permuting = 4 & \begin{tabular}[c]{@{}c@{}}{[}{[}{[}1, 2{]}, {[}3, 5, 4{]}, {[}6, 8, 9, 7{]}, {[}10, 11{]}, {[}12, 14, 13{]}{]},\\ {[}{[}1, 2{]}, {[}3, 5, 4{]}, {[}6, 9, 7, 8{]}, {[}10, 11{]}, {[}12, 14, 13{]}{]}{]}\end{tabular}                                                                                                                                                                                                                                                                                                                                                                                                                                                                                                                                                                                                                    \\ \hline
\end{tabular}
\end{table}

\subsection*{Appendix C.4: Experimental settings of Fig.~1 in the manuscript}
For ResNetx on CIFAR10 dataset, we directly utilize the pretrained models in a publicly available pytorch repository~\cite{Idelbayev18a}. This repository provides a valid implementation that matches with the description of the original paper~\cite{he2016deep}, with comparable or smaller test error. When testing, we always employ batchnorm re-calibration for sampled sub-networks following~\cite{yu2019universally}.

\section*{Appendix D: The trajectory of training loss and test accuracy}
\begin{figure}[H]
\vspace{-4mm}
  \centering
  \subfigure[MBV3 on C10]{
    \label{fig:subfig:onefunction} 
    \includegraphics[width=0.31\linewidth]{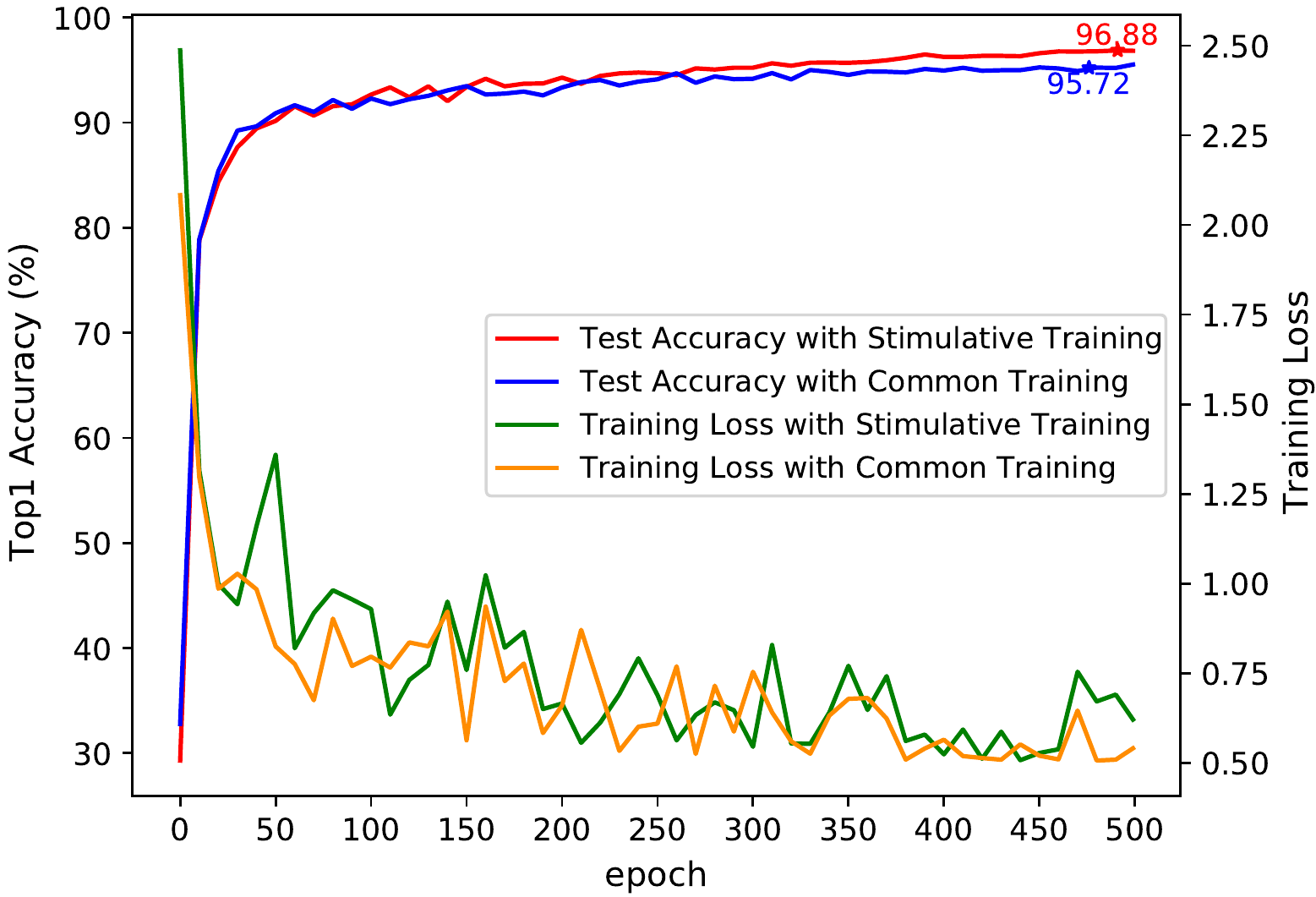}}
  \subfigure[MBV3 on C100]{
    \label{fig:subfig:twofunction} 
    \includegraphics[width=0.31\linewidth]{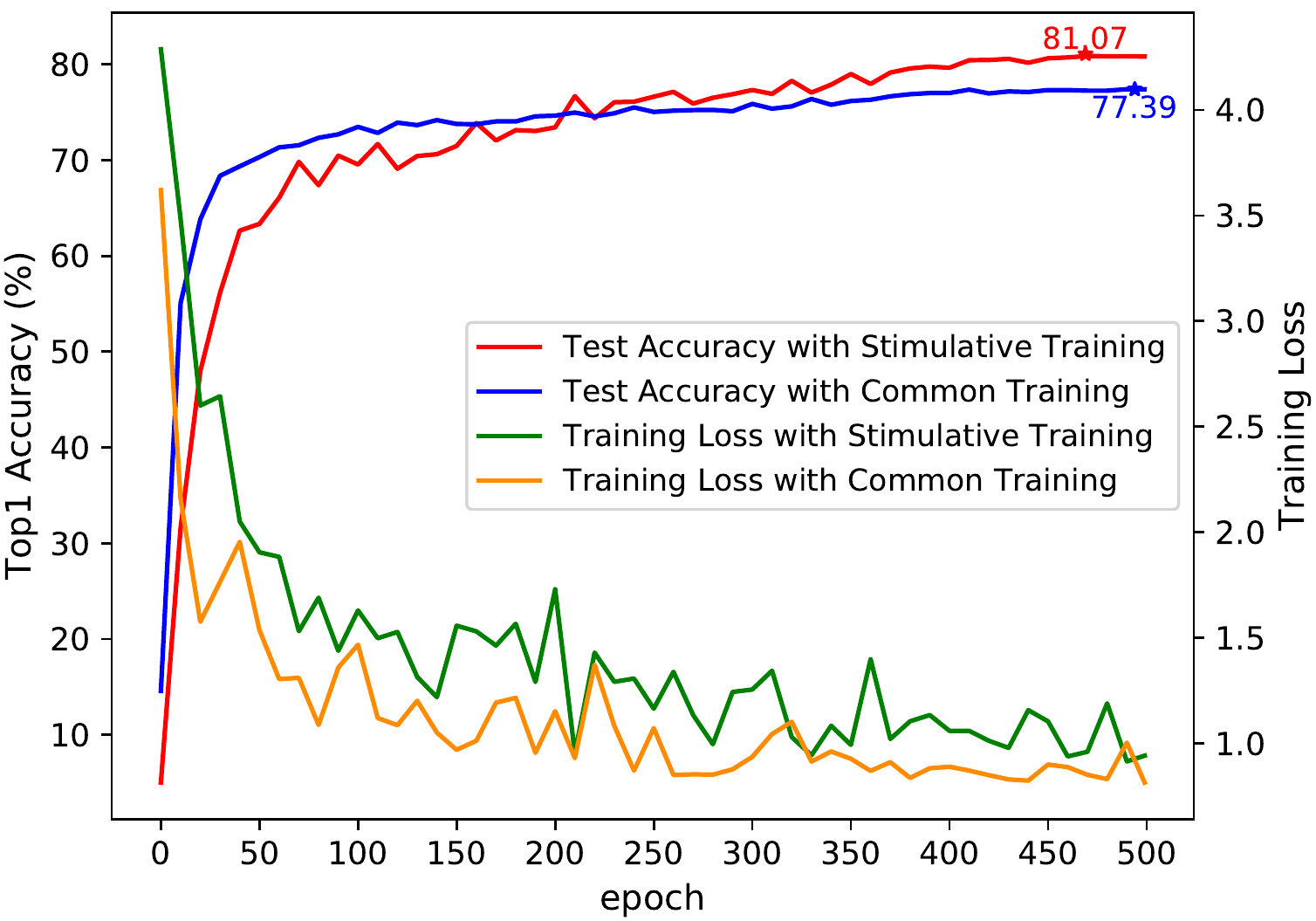}}
  \subfigure[Res50 on C100]{
    \label{fig:subfig:twofunction} 
    \includegraphics[width=0.31\linewidth]{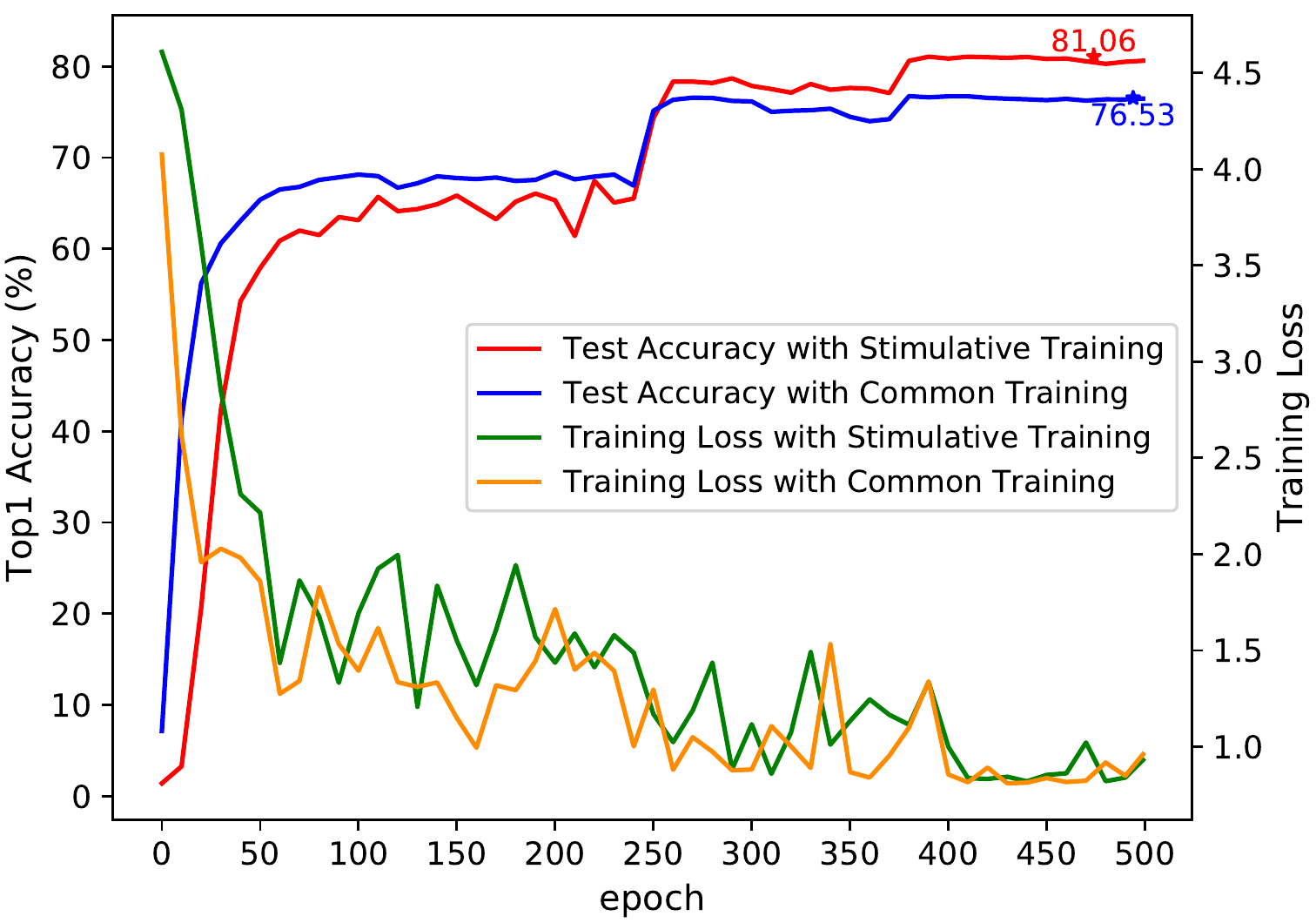}}
  \vspace{-4mm}
  \caption{The trajectory of training loss and test accuracy when applying stimulative and common training on different residual networks and benchmark datasets.}
  \label{fig:loss acc} 
\vspace{-4mm}
\end{figure}

We show the trajectory of training loss and test accuracy when applying stimulative and common training in Fig.~\ref{fig:loss acc}. The point of the highest Top1 accuracy is represented by five-pointed star for both approaches. As we can see, on different residual networks and benchmark datasets, stimulative training can always yield higher accuracy than common training (i.e., 96.88\% vs 95.72\%, 81.07\% vs 77.39\%, 81.06\% vs 76.53\%). Furthermore, we observe that the curves of Top1 accuracy and training loss of residual networks with common training rapidly approach flat, while stimulative training continuously reduces the training loss and increases the Top1 accuracy.

\section*{Appendix E: Ablation study of the balanced coefficient $\lambda$}
\begin{figure}[H]
\vspace{-4mm}
  \centering
  \subfigure[MBV3 on C10]{
    \label{fig:subfig:onefunction} 
    \includegraphics[width=0.31\linewidth]{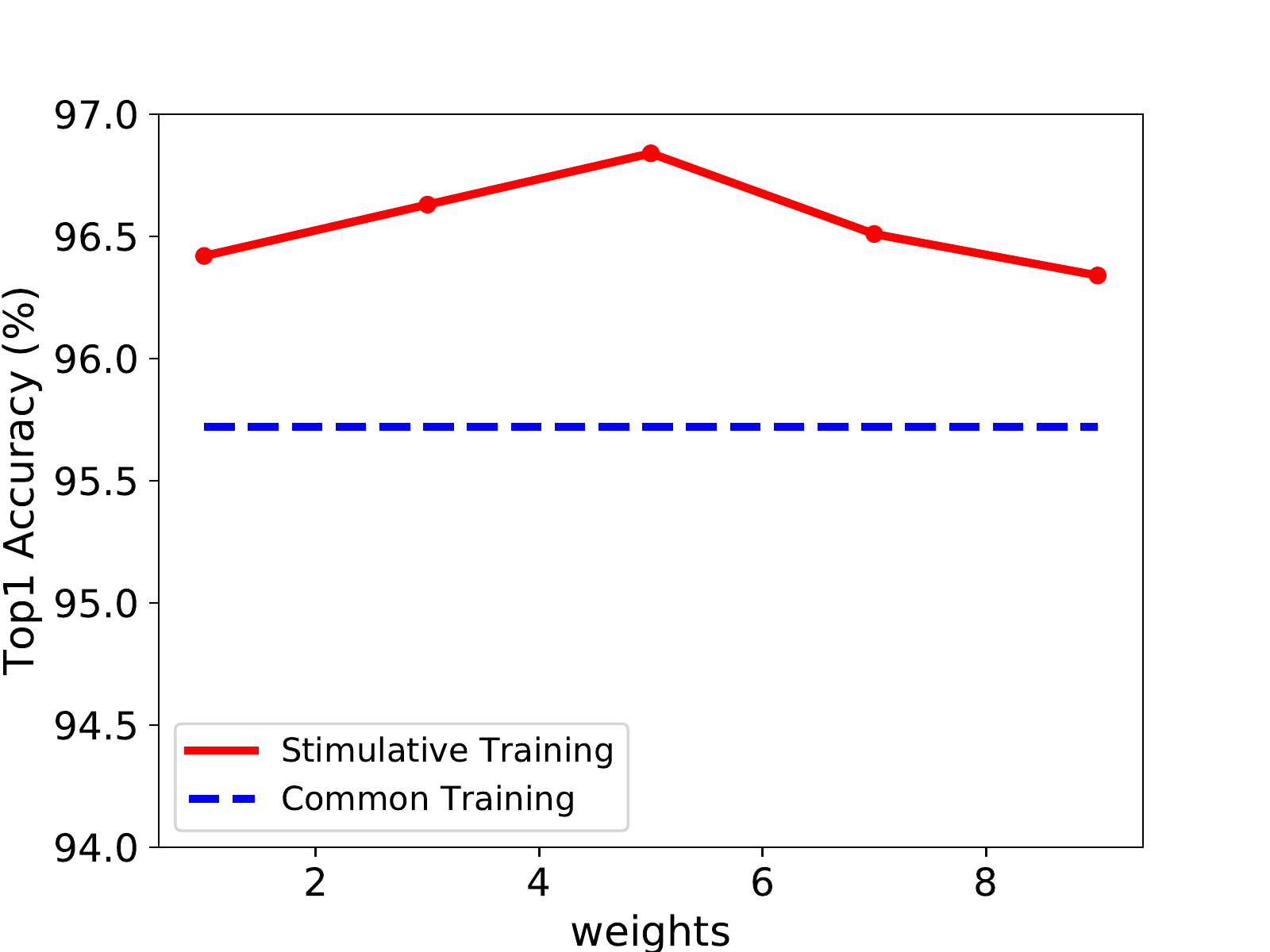}}
  \subfigure[MBV3 on C100]{
    \label{fig:subfig:twofunction} 
    \includegraphics[width=0.31\linewidth]{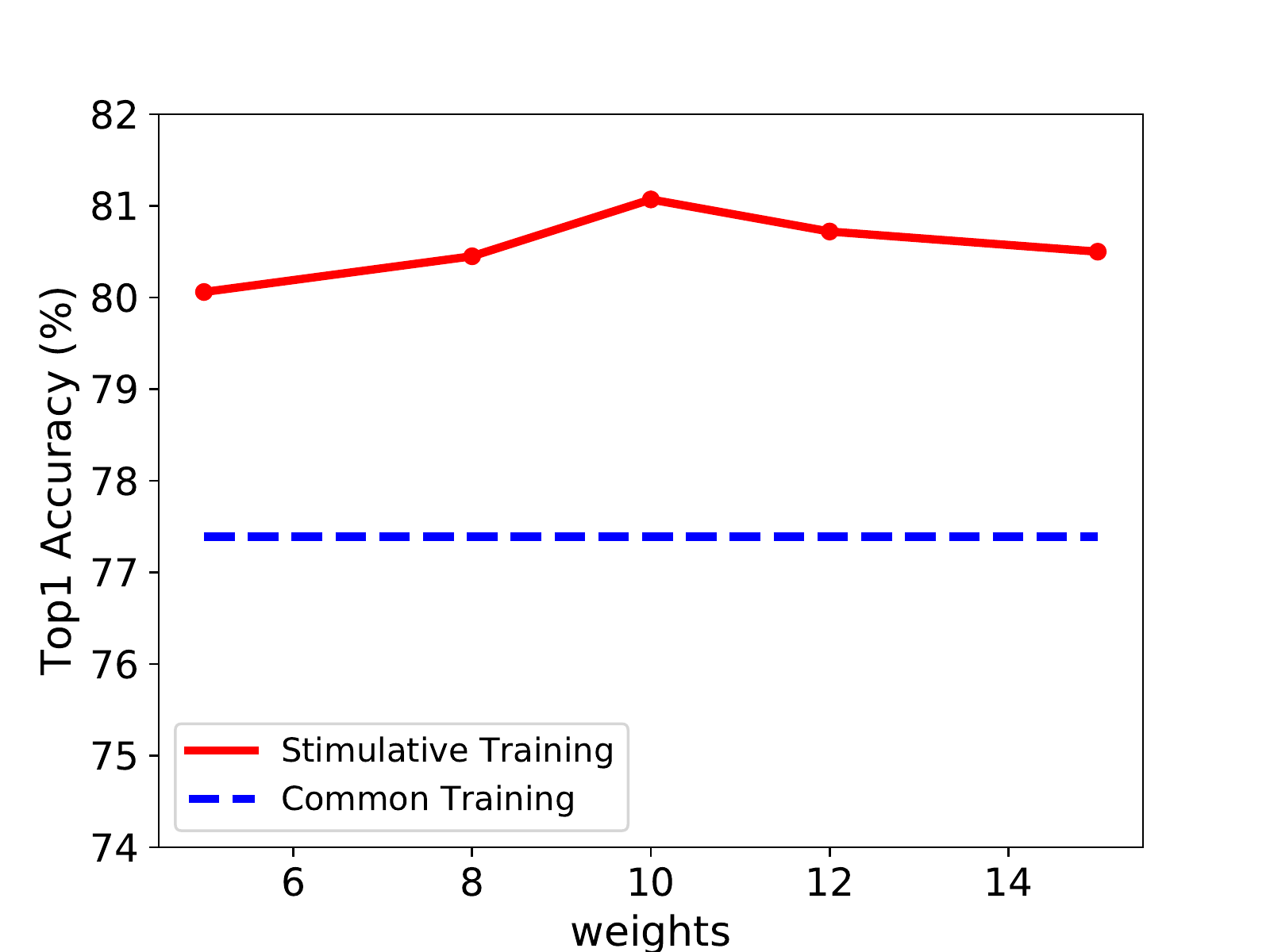}}
  \subfigure[Res50 on C100]{
    \label{fig:subfig:twofunction} 
    \includegraphics[width=0.31\linewidth]{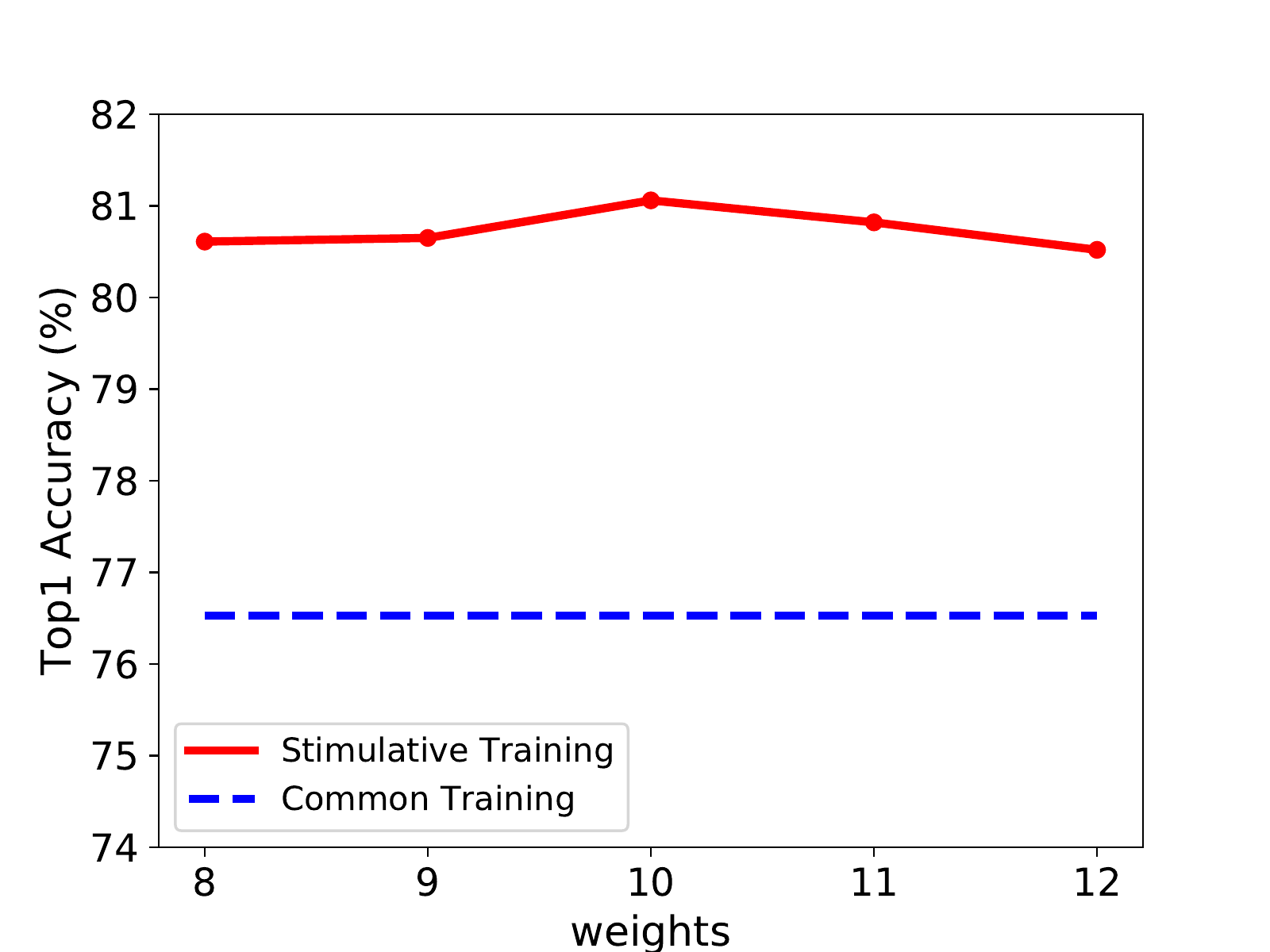}}
  \vspace{-4mm}
  \caption{Ablation study of the balanced coefficient $\lambda$ between cross entropy loss and KL divergence loss on different residual networks and benchmark datasets.}
  \label{fig:coefficient} 
\vspace{-4mm}
\end{figure}

We show the ablation study of the balanced coefficient $\lambda$ between cross entropy loss and KL divergence loss in Fig.~\ref{fig:coefficient}. As we can see, on different residual networks and benchmark datasets, stimulation training with different balance coefficients may yield slightly different Top1 accuracies, but all these results are much better than those obtained by common training. In addition, as shown in Fig.~\ref{fig:coefficient}, the optimal balance coefficients for MobileNetV3 on CIFAR10, MobileNetV3 on CIFAR100 and ResNet50 on CIFAR100 are 5, 10 and 10 respectively. Similar results can be found for ResNet families on ImageNet, and we find that their optimal balance coefficients are 1.

\section*{Appendix F: Rebuttle additional part}
 
 \subsection*{Appendix F.1: Comparison with different methods}
We further compare the proposed stimulative training with self distillation~\cite{zhang2019your}, stochastic depth~\cite{huang2016deep}, and common training with providing supervision directly to each layer or stage. We train MobileNetV3 with these methods on CIFAR100 dataset. For a fair comparison, we adopt the same basic training settings such as training epoch, data augmentation, learning rate and optimizer, which has been shown in the manuscript Section.6. The detailed respective training settings are given as follows.
 \par For self distillation we adopt the similar hyper-parameters as~\cite{zhang2019your} and fine tune them for MobileNetV3. In the logits distillation, the main loss coefficient is 0.8 and distillation coefficient is 0.2. In the feature distillation, the feature distillation coefficient of each stage is 0.02.
 \par For stochastic depth, due to forbidding skipping down sampling layer in MobileNetV3, we ensure the first layer of each stage won't be dropped. The survive rate of layers are declined linearly~\cite{huang2016deep} and the final survive rate p=0.9.
 \par For layer and stage supervision training, we introduce additional transforming heads to project the features of stages or layers to the same dimension space and utilize cross entropy to generate supervision. The stage feature loss coefficient is [0.13, 0.2, 0.27, 0.4] and layer feature loss coefficient is the corresponding stage coefficient divided by the layer number of stage.
 \par
\begin{figure}[H]
\vspace{-4mm}
  \centering
  \subfigure[deleting one]{
    \label{fig:subfig:sd_deleting one} 
    \includegraphics[width=0.31\linewidth]{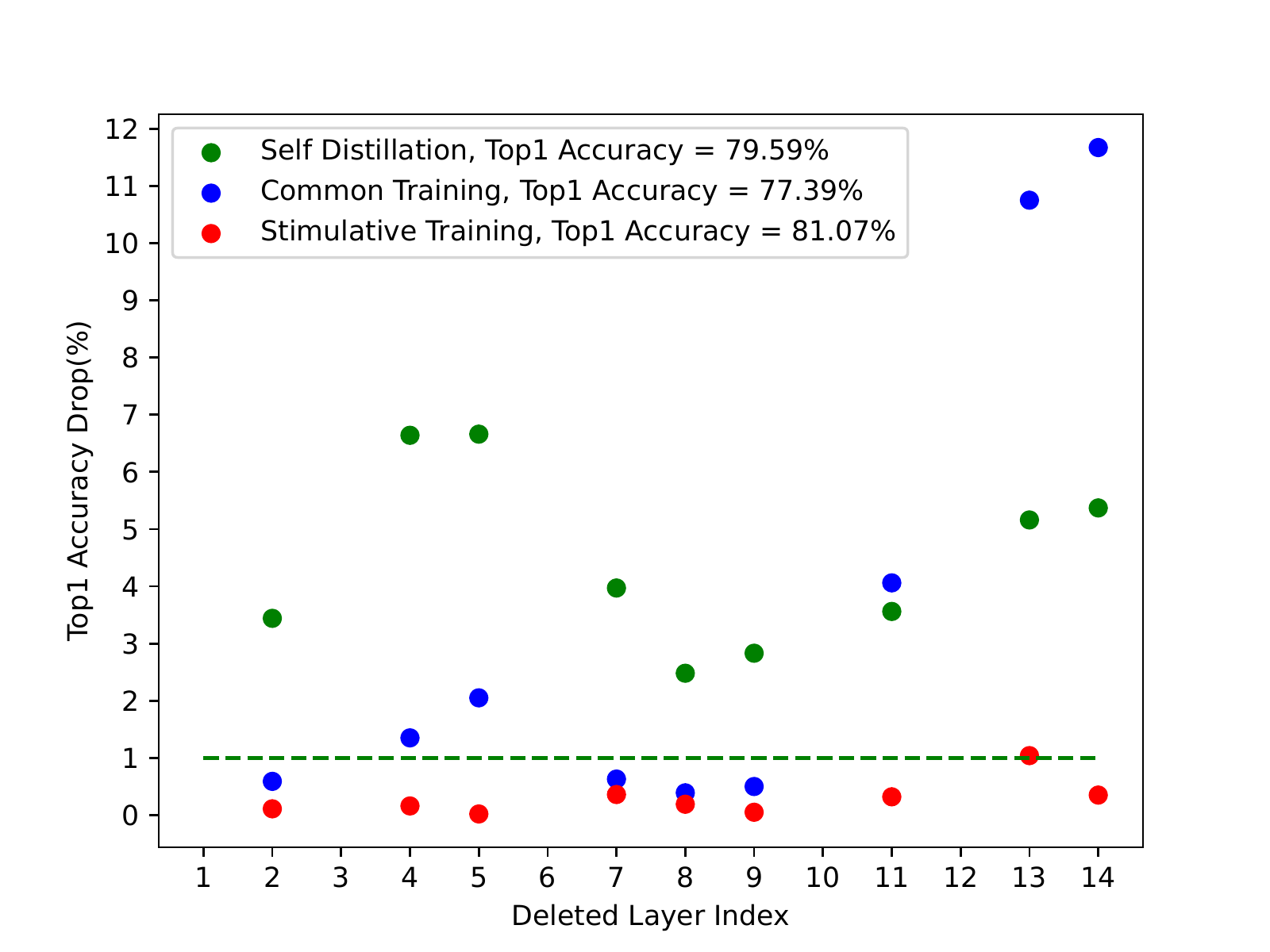}}
  \subfigure[deleting more]{
    \label{fig:subfig:sd_deleting more} 
    \includegraphics[width=0.31\linewidth]{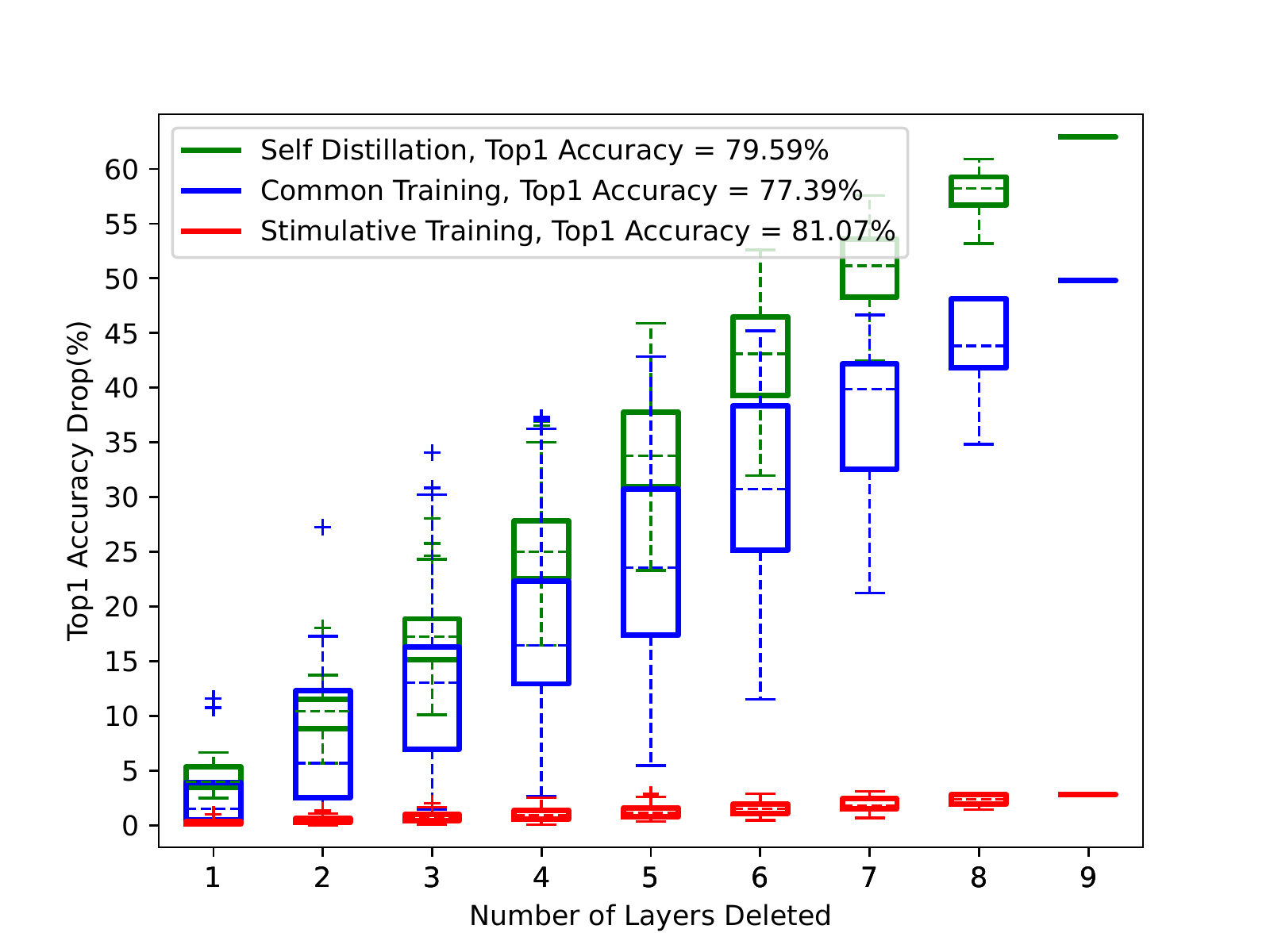}}
  \subfigure[permuting]{
    \label{fig:subfig:sd_permuting} 
    \includegraphics[width=0.31\linewidth]{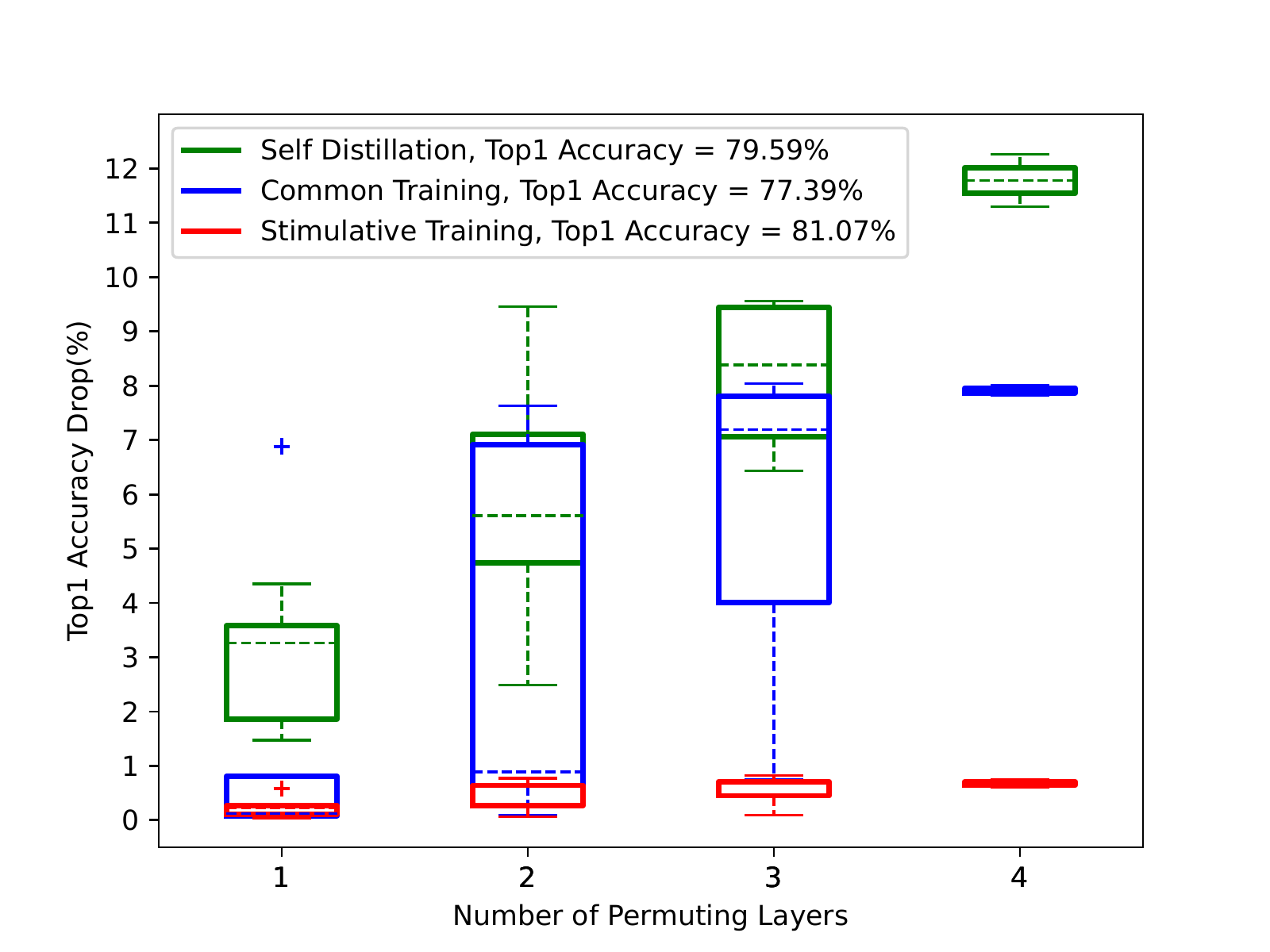}}
  \vspace{-4mm}
  \caption{Top1 accuracy drop when (a) deleting one layer, (b) deleting more layers and (c) permuting
layers from MobileNetV3 with stimulative/common/self distillation training on CIFAR100 dataset.}
  \label{fig:sd fig}
\vspace{-4mm}
\end{figure}

\begin{figure}[H]
\vspace{-4mm}
  \centering
  \subfigure[deleting one]{
    \label{fig:subfig:StoD_deleting one} 
    \includegraphics[width=0.31\linewidth]{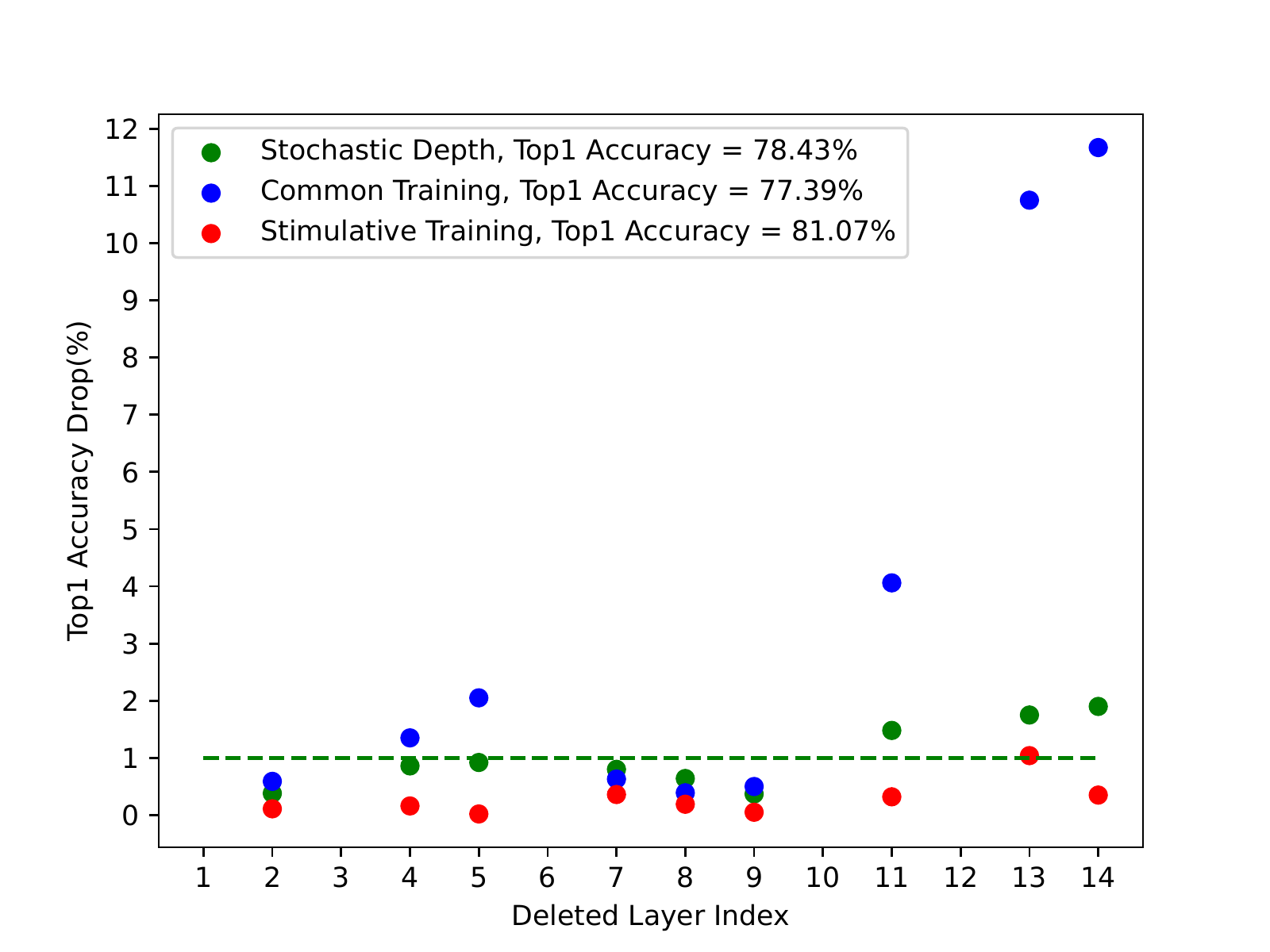}}
  \subfigure[deleting more]{
    \label{fig:subfig:StoD_deleting more} 
    \includegraphics[width=0.31\linewidth]{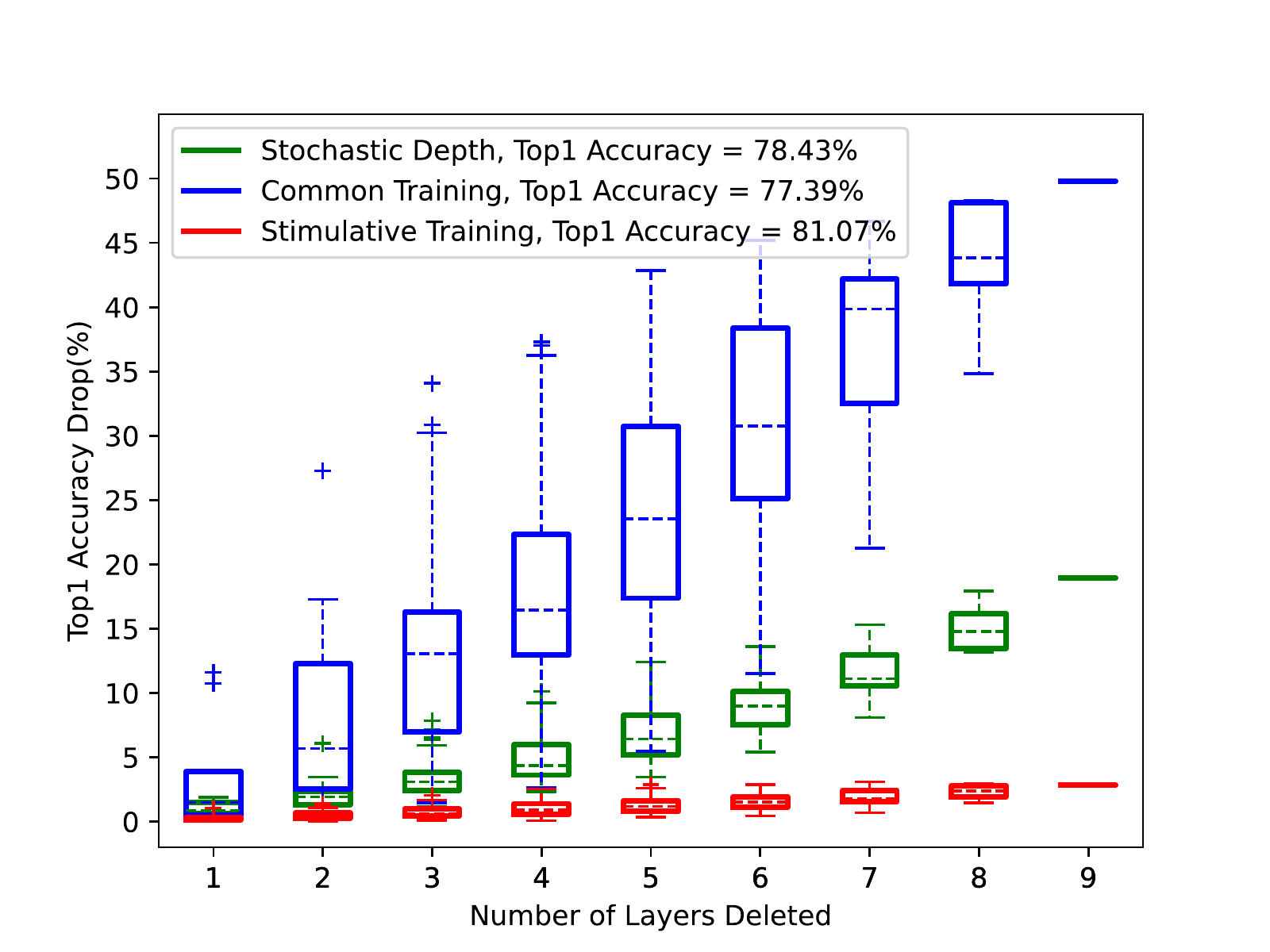}}
  \subfigure[permuting]{
    \label{fig:subfig:StoD_permuting} 
    \includegraphics[width=0.31\linewidth]{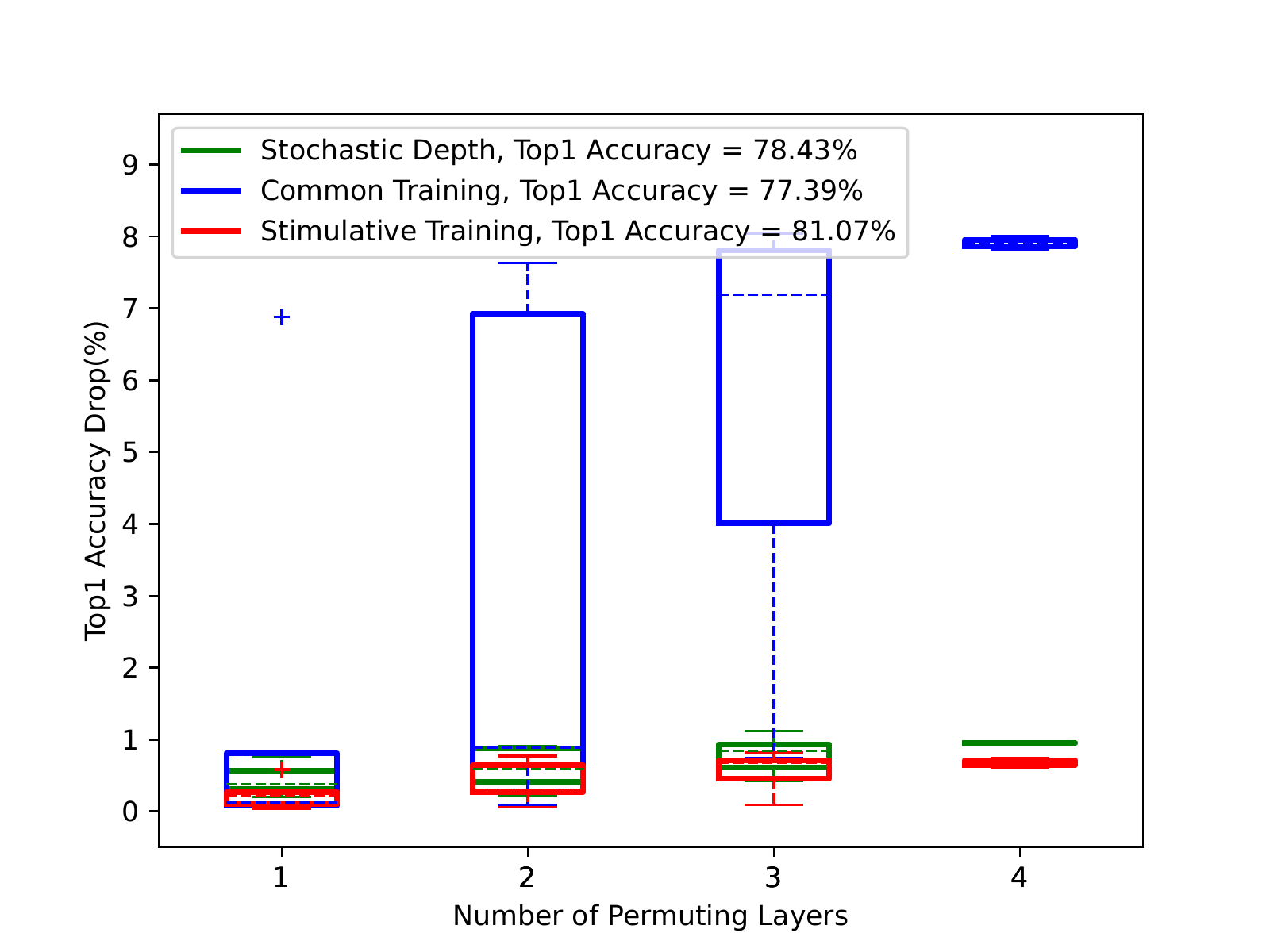}}
  \vspace{-4mm}
  \caption{Top1 accuracy drop when (a) deleting one layer, (b) deleting more layers and (c) permuting
layers from MobileNetV3 with stimulative/common/stochastic depth training on CIFAR100 dataset.}
  \label{fig:StoD fig}
\vspace{-4mm}
\end{figure}

\begin{figure}[H]
\vspace{-4mm}
  \centering
  \subfigure[deleting one]{
    \label{fig:subfig:sl_deleting one} 
    \includegraphics[width=0.31\linewidth]{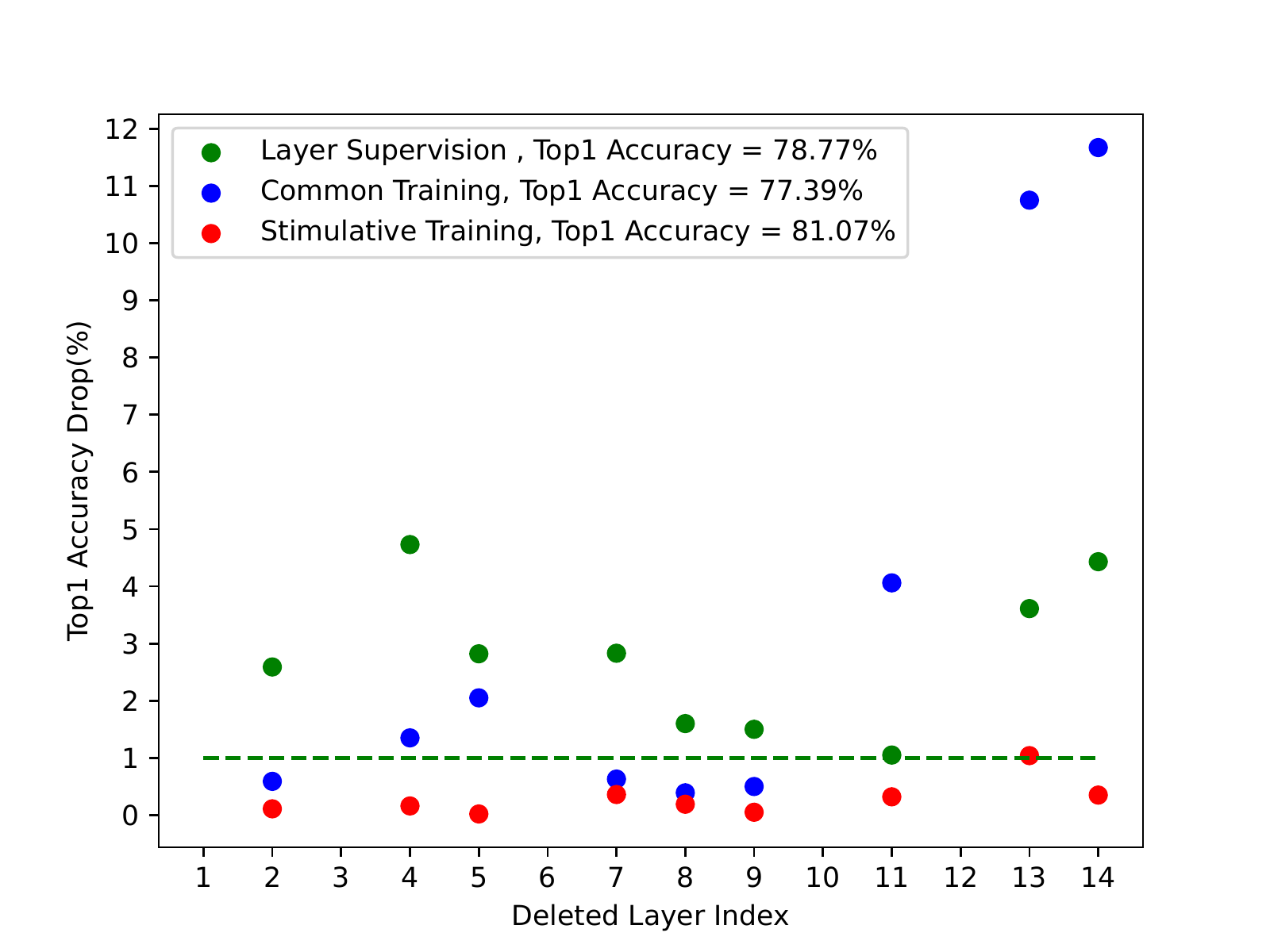}}
  \subfigure[deleting more]{
    \label{fig:subfig:sl_deleting more} 
    \includegraphics[width=0.31\linewidth]{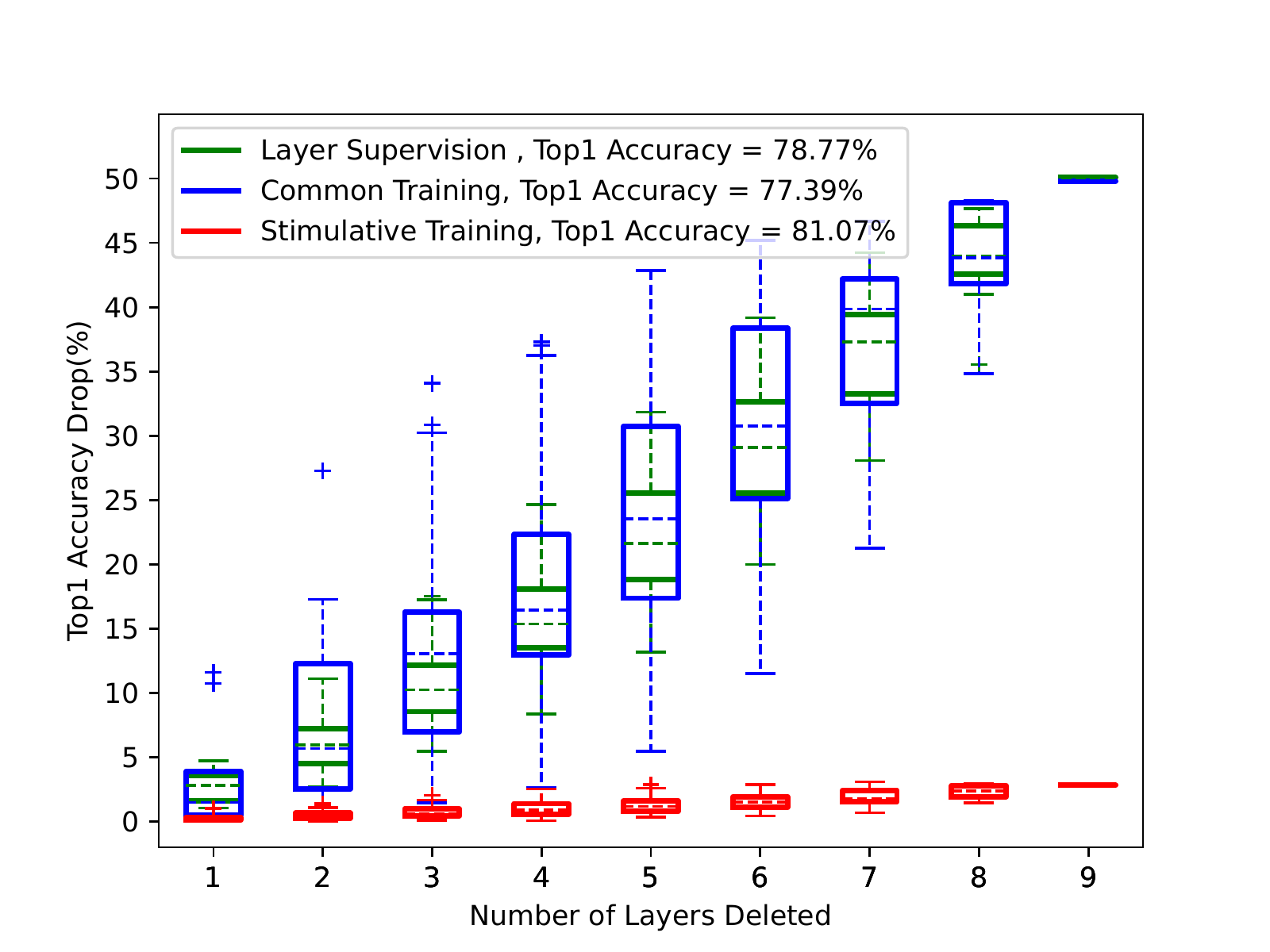}}
  \subfigure[permuting]{
    \label{fig:subfig:sl_permuting} 
    \includegraphics[width=0.31\linewidth]{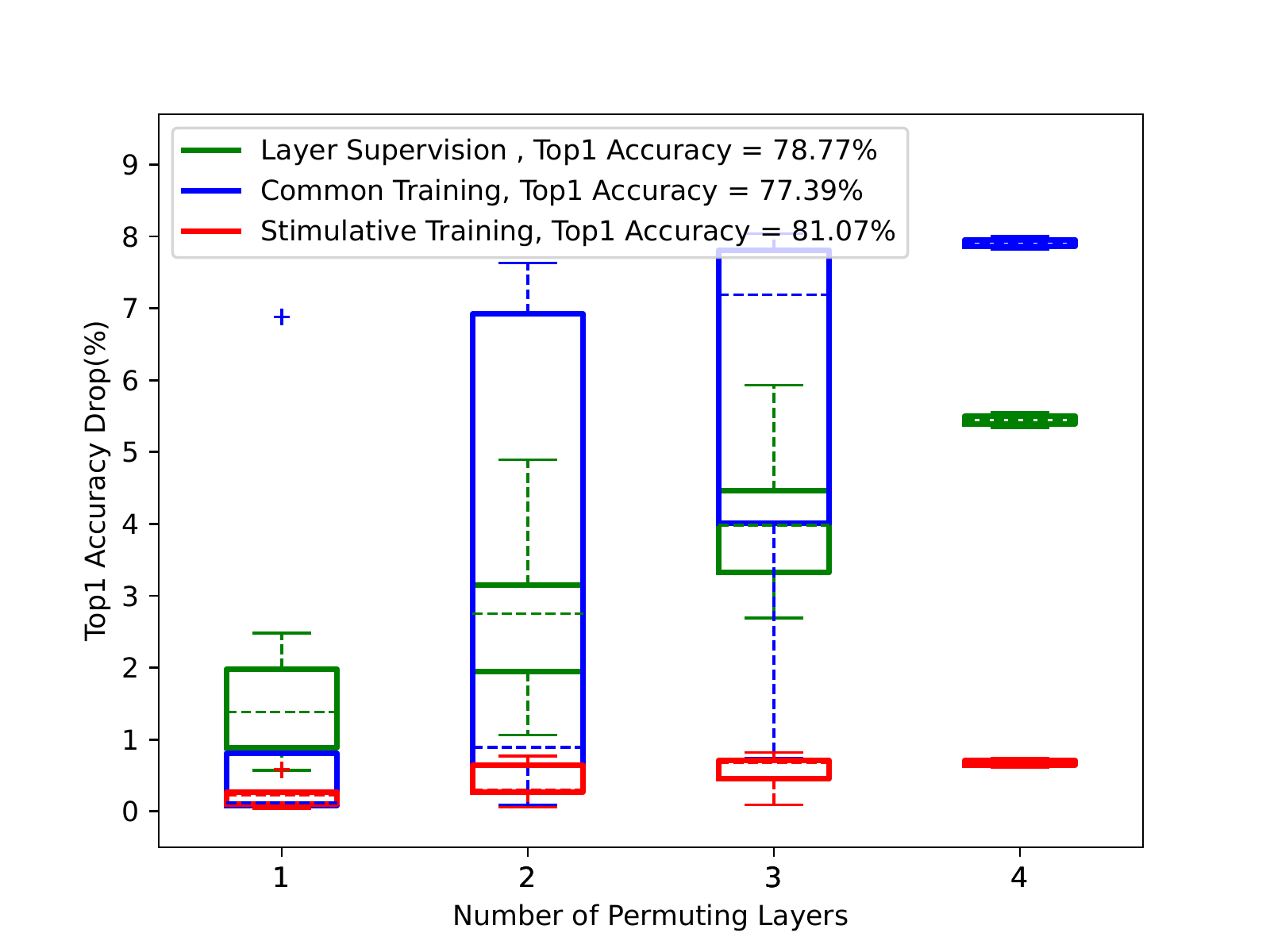}}
  \vspace{-4mm}
  \caption{Top1 accuracy drop when (a) deleting one layer, (b) deleting more layers and (c) permuting
layers from MobileNetV3 with stimulative/common/layer supervision training on CIFAR100 dataset.}
  \label{fig:supervision layer fig}
\vspace{-4mm}
\end{figure}

\begin{figure}[H]
\vspace{-4mm}
  \centering
  \subfigure[deleting one]{
    \label{fig:subfig:ss_deleting one} 
    \includegraphics[width=0.31\linewidth]{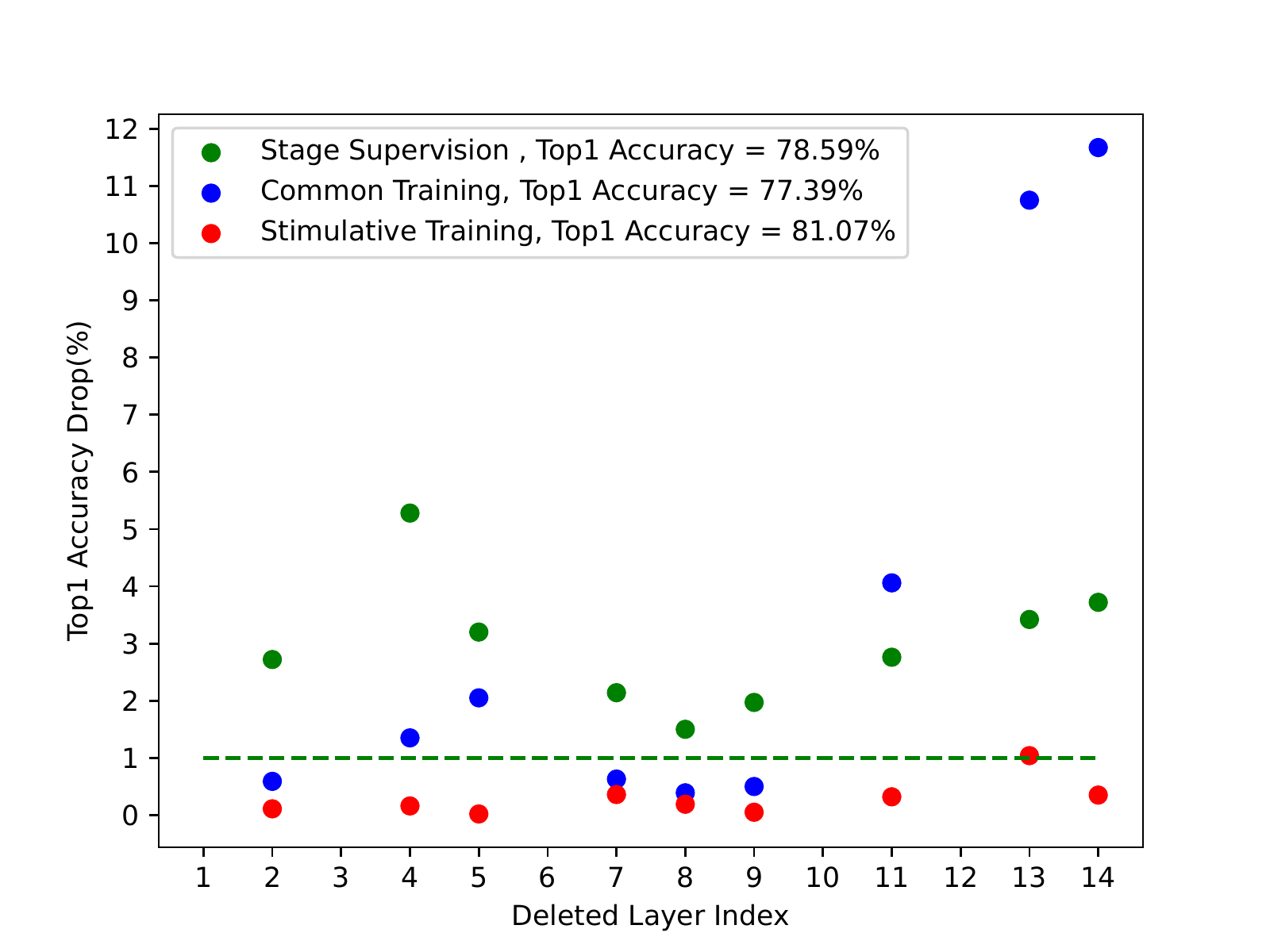}}
  \subfigure[deleting more]{
    \label{fig:subfig:ss_deleting more} 
    \includegraphics[width=0.31\linewidth]{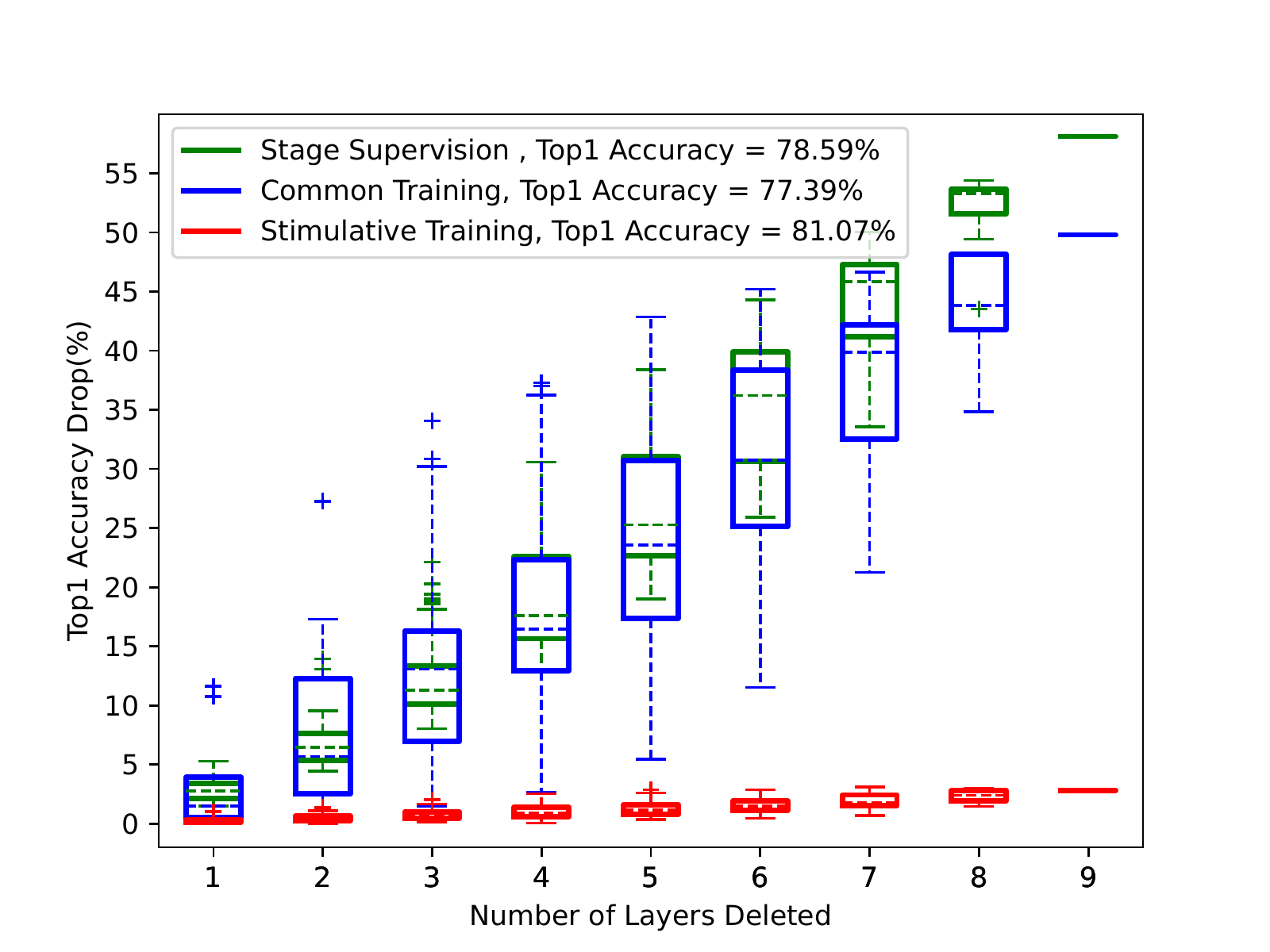}}
  \subfigure[permuting]{
    \label{fig:subfig:ss_permuting} 
    \includegraphics[width=0.31\linewidth]{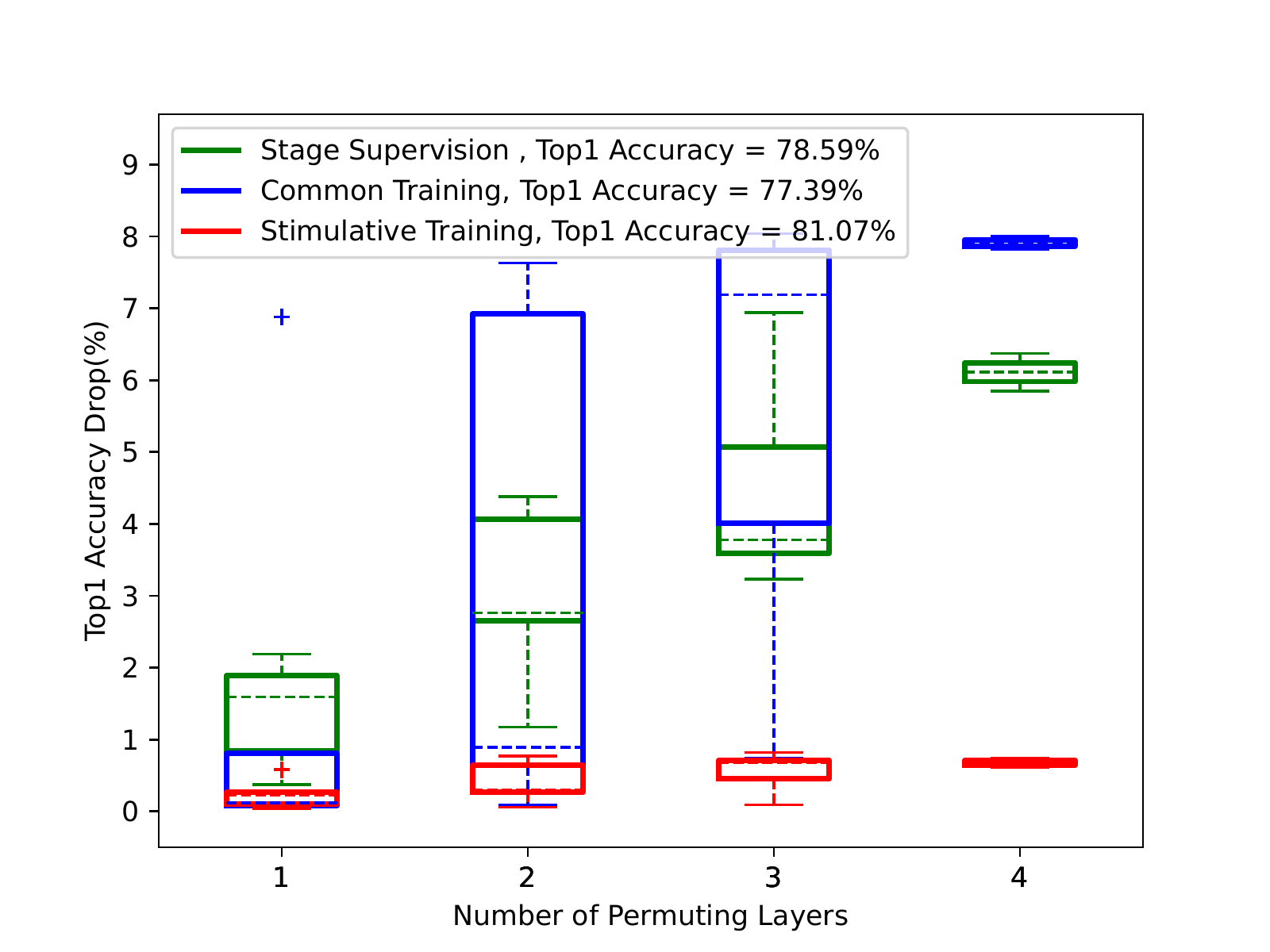}}
  \vspace{-4mm}
  \caption{Top1 accuracy drop when (a) deleting one layer, (b) deleting more layers and (c) permuting
layers from MobileNetV3 with stimulative/common/stage supervision training on CIFAR100 dataset.}
  \label{fig:supervision stage fig}
\vspace{-4mm}
\end{figure}

\begin{figure}[H]
\centering
\subfigure[Self distillation]{\label{fig:subfig:SD_accuracy}
\includegraphics[width=0.31\linewidth]{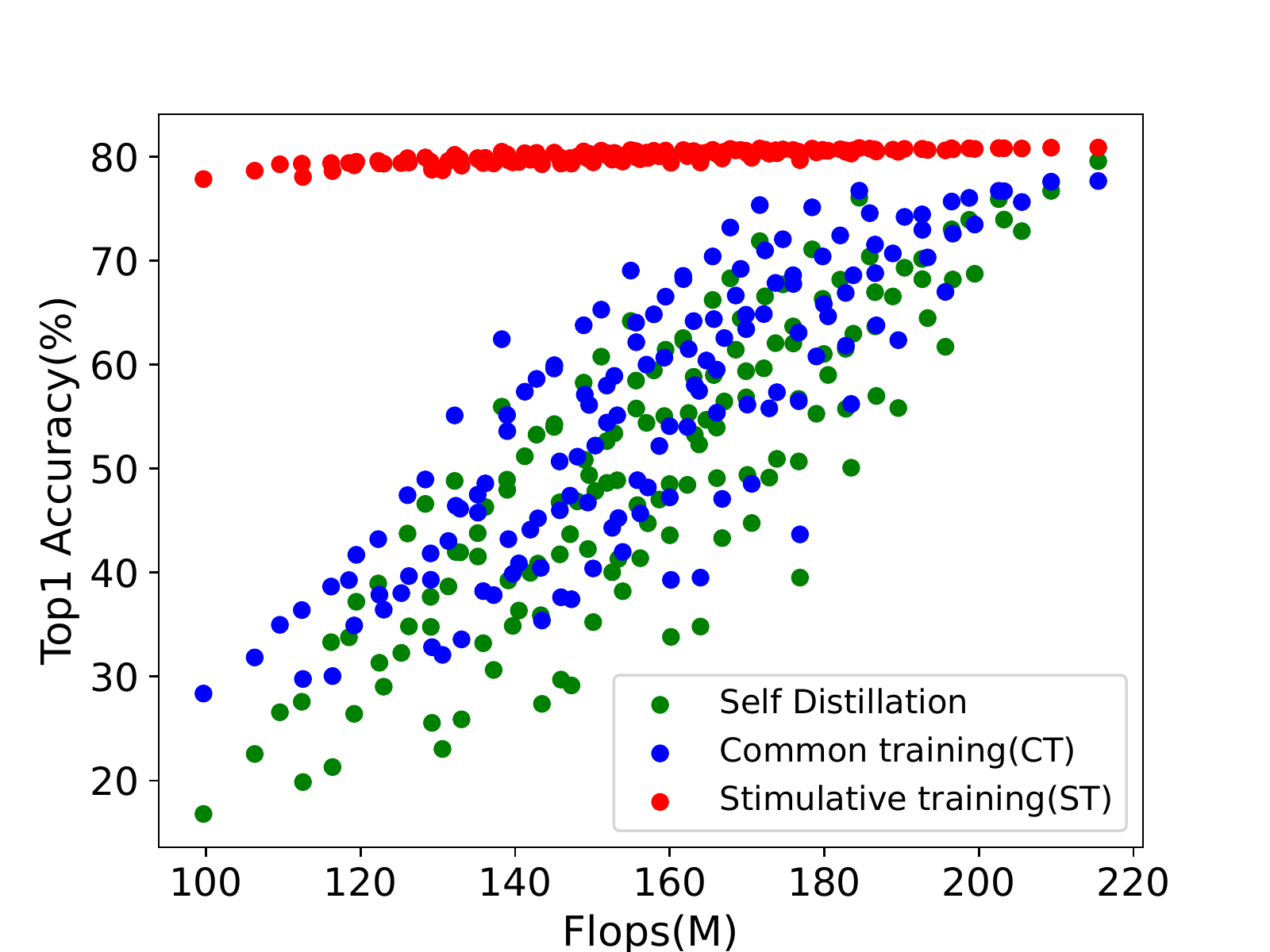}}
\hspace{0.01\linewidth}
\subfigure[Stochastic depth]{\label{fig:subfig:StoD_accuracy}
\includegraphics[width=0.31\linewidth]{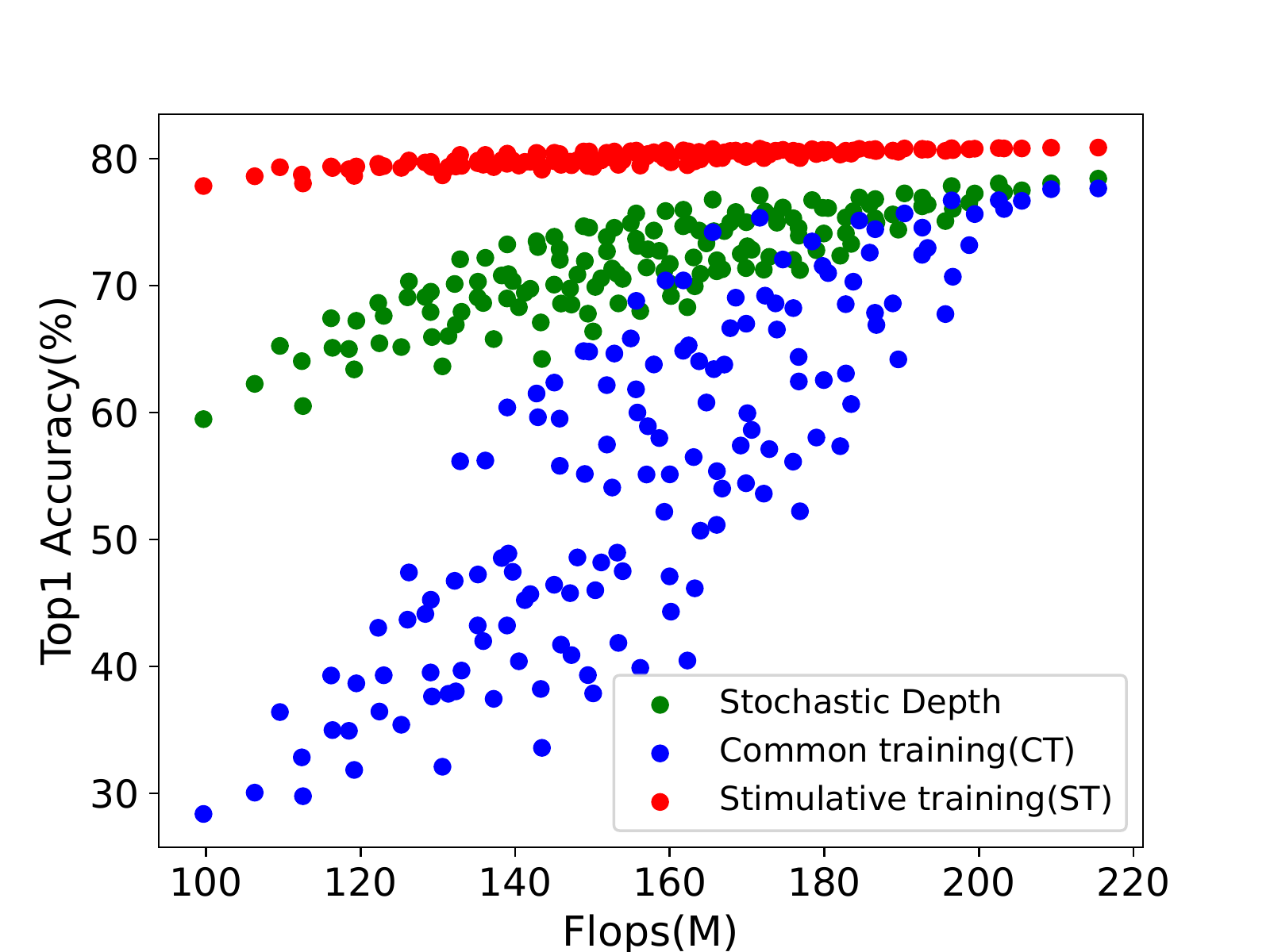}}
\vfill
\subfigure[Layer supervision]{\label{fig:subfig:ls_accuracy}
\includegraphics[width=0.31\linewidth]{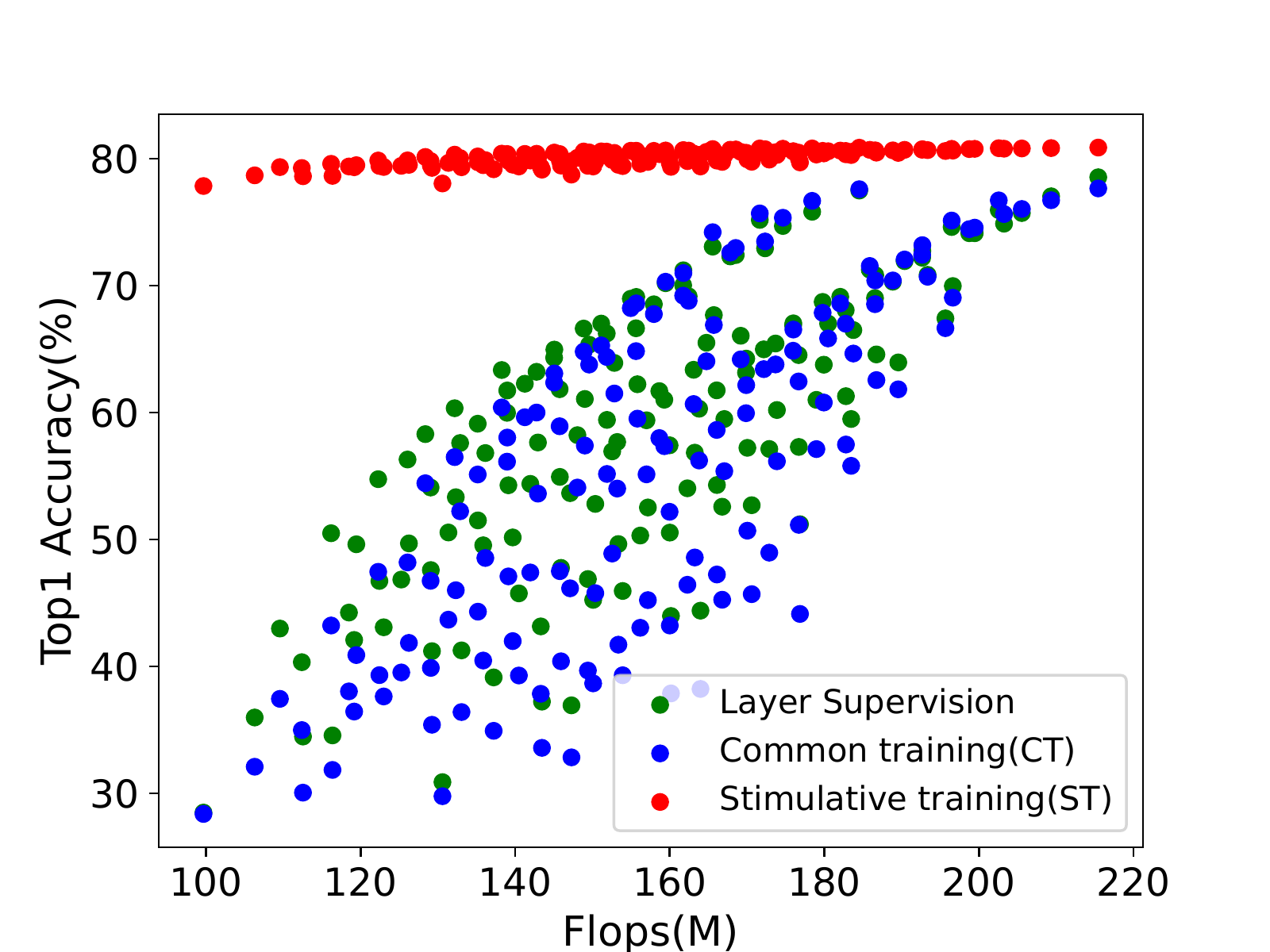}}
\hspace{0.01\linewidth}
\subfigure[Stage supervision]{\label{fig:subfig:ss_accuracy}
\includegraphics[width=0.31\linewidth]{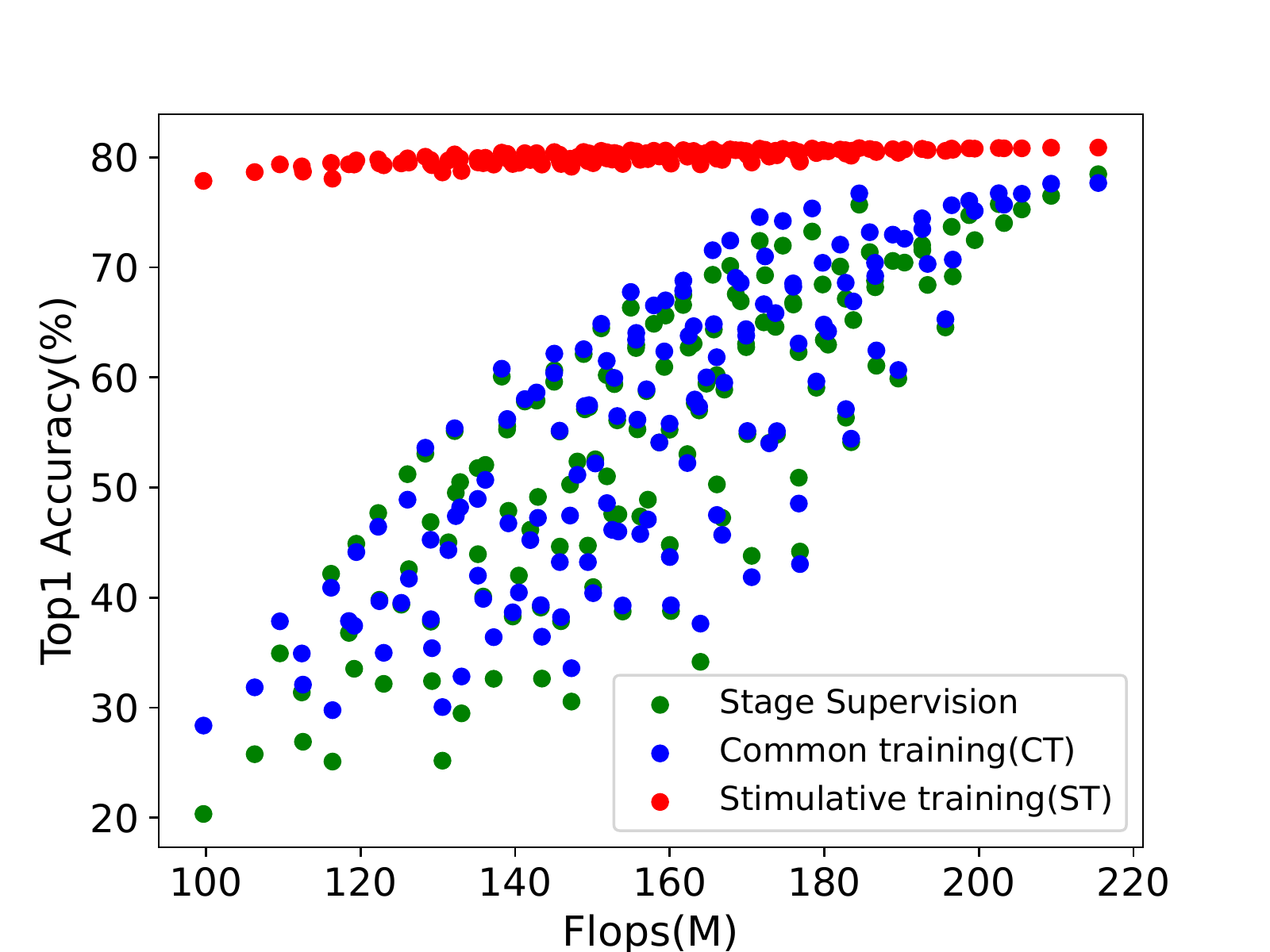}}
\caption{Comparisons of sub-networks performance on MBV3 with different training methods. As we can see, the proposed stimulative training can better relieve the network loafing problem than all the other methods.}
\label{fig:Accuracy compare}
\end{figure}

 The comprehensive comparisons are shown in Table~\ref{table:performance of network}.As we can see, layer supervision and stochastic depth can improve both the performance of the main network and the average performance of all subnetworks, stage supervision and self-distillation can only improve the performance of the main network, while the proposed stimulative training can achieve the highest performance of main network and the highest average performance of all subnetworks. Besides, as shown in Fig~\ref{fig:Accuracy compare}, the proposed stimulative training can better relieve the network loafing problem than all the other methods. As shown in Fig~\ref{fig:sd fig}, Fig~\ref{fig:StoD fig}, Fig~\ref{fig:supervision layer fig} and Fig~\ref{fig:supervision stage fig}, the proposed stimulative training can provide stronger robustness in resisting various network destruction operations than all the other methods. 
 
 Besides above experimental results, we find that: 1) The improved performance of stochastic depth can be also interpreted as relieving the loafing problem defined in this work; 2) the proposed stimulative training is actually complementary to layer/stage supervision and self-distillation, and their combinations can be a worthy research direction in the future.
 
 \subsection*{Appendix F.2: Loafing of DenseNet networks}
 To show that the loafing problem are replicable, we further verify that DenseNet also suffers from the loafing problem on ImageNet and CIFAR100. We follow \cite{huang2017densely} to select 4 typical networks (DenseNet121, 169, 201, 264), in which the tinier one is completely the sub-network of the larger one, to validate the loafing problem. For ImageNet, we just apply the pretrained weights of DenseNet121, 169, 201 and test the sub-network from larger one by succeeding the weights. For CIFAR100, we train all 4 networks utilizing standard training setting and do the same test above. For details, the optimizer is SGD and the initial learning rate is 0.1 and divided by 5 at 60th, 120th, 160th for 200 epochs with batchsize 128, weight decay 5e-4, Nesterov momentum of 0.9. 
 \par Table \ref{table:densenet imagenet} and Table \ref{table:densenet cifar100} show the results of different DenseNet networks which are trained on ImageNet and CIFAR100 respectively. As we can see that, different DenseNet networks invariably suffer from the loafing problem, that is, the sub-networks working in a given DenseNet network are prone to exert degraded performances than these sub-networks working individually. Moreover, the loafing problem of deeper DenseNet networks is inclined to be more severe than that of shallower ones, that is, the same sub-network in deeper DenseNet networks constantly presents inferior performance than that in shallower DenseNet networks. As Fig 1 of the main text has shown that different ResNet networks invariably suffer from the loafing problem, we can conclude that various residual networks invariably suffer from the loafing problem on various datasets.
 
 \subsection*{Appendix F.3: Training time and memory cost}
 We further show the difference in training time and memory consumption between common training (CT) and stimulative training (ST) in Table~\ref{table:time and memory}. For training time, ST is about 1.4 times (but less than 2 times) that of CT, since ST employs the main network and a sampled subnetwork at each step and the sampled subnetwork usually takes much less time than the main network. For memory consumption, ST and CT are basically the same, since each subnetwork is sampled from the main network.
 



\begin{table}[H]
\vspace{-2mm}
\caption{\centering Comprehensive comparisons among different methods, including common training (CT), stimulative training (ST), common training (CT) with layer/stage supervision, Self-Distillation and Stochastic Depth.}
\vspace{-2mm}
\centering
\label{table:performance of network}
\begin{tabular}{|l|l|l|l|l|}
\hline
Method               & Time           & Memory           & Main(\%)       & All(\%)             \\ \hline
CT                   & 16.91h         & 3291MiB          & 77.39          & 55.26±13.37         \\ \hline
CT+layer supervision & 23.3h          & 7193MiB          & 78.77          & 59.18±11.12         \\ \hline
CT+stage supervision & 19.3h          & 5197MiB          & 78.59          & 54.82±13.31         \\ \hline
Self distillation    & 26.8h          & 3887MiB          & 79.59          & 50.39±14.22         \\ \hline
Stochastic depth     & \textbf{16.9h} & \textbf{3291MiB} & 78.43          & 70.72±3.76          \\ \hline
ST                   & 24.08h         & \textbf{3291MiB}          & \textbf{81.07} & \textbf{80.01±0.59} \\ \hline
\end{tabular}
\vspace{-4mm}
\end{table}

\begin{table}[H]
\vspace{-2mm}
\caption{\centering Training time and memory consumption on different models and datasets. CT and ST denote common training and stimulative training, respectively.}
\vspace{-2mm}
\centering
\label{table:time and memory}
\begin{tabular}{|l|l|l|l|}
\hline
Method      & MBV3\_C10 & MBV3\_C100 & Res50\_C100 \\ \hline
CT (time)   & 16.77h    & 16.91h     & 15.28h      \\ \hline
ST (time)   & 23.64h    & 24.08h     & 21.52h      \\ \hline
CT (memory) & 3361MiB   & 3291MiB    & 4647MiB     \\ \hline
ST (memory) & 3361MiB   & 3291MiB    & 4647MiB     \\ \hline
\end{tabular}
\vspace{-4mm}
\end{table}

\begin{table}[H]
\vspace{-2mm}

\caption{\centering Different DenseNet networks trained on ImageNet invariably suffer from the problem of network loafing.}
\vspace{-2mm}
\centering
\label{table:densenet imagenet}
\begin{tabular}{|l|l|l|l|}
\hline
Main-net\textbackslash{}Sub-net & DenseNet121 & DenseNet169 & DenseNet201 \\ \hline
DenseNet121                     & 74.86       & 20.91       & 11.57       \\ \hline
DenseNet169                     & -           & 76.46       & 51.18       \\ \hline
DenseNet201                     & -           & -           & 77.44       \\ \hline
\end{tabular}
\vspace{-4mm}
\end{table}

\begin{table}[H]
\vspace{-2mm}
\caption{\centering Different DenseNet networks trained on CIFAR100 invariably suffer from the problem of network loafing.}
\vspace{-2mm}
\centering
\label{table:densenet cifar100}
\begin{tabular}{|l|l|l|l|l|}
\hline
Main-net\textbackslash{}Sub-net & DenseNet121 & DenseNet169 & DenseNet201 & DenseNet264 \\ \hline
DenseNet121                     & 78.84       & 43.64       & 31.51       & 10.01       \\ \hline
DenseNet169                     & -           & 79.64       & 70.78       & 48.41       \\ \hline
DenseNet201                     & -           & -           & 79.77       & 62.29       \\ \hline
DenseNet264                     & -           & -           & -           & 79.81       \\ \hline
\end{tabular}
\vspace{-4mm}
\end{table}

\bibliography{neurips_2022}